\documentclass[10pt,twocolumn,letterpaper]{article}

\usepackage{cvpr}
\usepackage{times}
\usepackage{epsfig}
\usepackage{graphicx}
\usepackage{amsmath}
\usepackage{amssymb}
\usepackage{color}

\usepackage{booktabs}       
\usepackage{paralist}
\usepackage{multirow}

\usepackage[pagebackref=true,breaklinks=true,letterpaper=true,colorlinks,bookmarks=false,citecolor=blue,linkcolor=blue]{hyperref}

\cvprfinalcopy 

\def\cvprPaperID{1435} 

\ifcvprfinal\fi
\begin{document}
	
\title{Learning to Adapt Structured Output Space for Semantic Segmentation}
	\author{
		Yi-Hsuan Tsai$^{1}$\thanks{Both authors contribute equally to this work.}
		\hspace{0.15in} Wei-Chih Hung$^{2\ast}$
		\hspace{0.15in} Samuel Schulter$^1$ 
		\hspace{0.15in} Kihyuk Sohn$^1$ \vspace{1mm}\\
		\hspace{0.15in} Ming-Hsuan Yang$^2$
		\hspace{0.15in} Manmohan Chandraker$^1$\vspace{1mm}\\
		\hspace{0.1in} $^1$NEC Laboratories America \hspace{0.15in} $^2$University of California, Merced
	}
	
	\maketitle
	
	\begin{abstract}
Convolutional neural network-based approaches for semantic segmentation rely on supervision with pixel-level ground truth, but
	may not generalize well to unseen image domains.
As the labeling process is tedious and labor intensive, developing algorithms that can adapt source ground truth labels to the target domain is of great interest.
In this paper, we propose an adversarial learning method for domain adaptation in the context of semantic segmentation.
Considering semantic segmentations as structured outputs that contain spatial similarities between the source and target domains, we adopt adversarial learning in the output space.
To further enhance the adapted model, we construct a multi-level adversarial network to effectively perform output space domain adaptation at different feature levels.
%
	%
Extensive experiments and ablation study are conducted under various domain adaptation settings, including synthetic-to-real and cross-city scenarios.
We show that the proposed method performs favorably against the state-of-the-art methods in terms of accuracy and visual quality.
	\end{abstract}

	\section{Introduction}
Semantic segmentation aims to assign each pixel a semantic label, e.g., person, car, road or tree, in an image. 
Recently, methods based on convolutional neural networks (CNNs)  
have achieved significant progress in semantic segmentation~\cite{deeplab, Lin_CVPR_2016, Liu_ICCV_2015, Long_CVPR_2015, Yu_ICLR_2016, pspnet, Zheng_ICCV_2015} with applications for autonomous driving~\cite{Geiger_CVPR_2012} and image editing~\cite{Tsai_CVPR_2017}.
The crux of CNN-based approaches is to annotate a large number of images that cover possible scene variations.  
However, this trained model may not generalize well to unseen images, especially when there is a domain gap between the training (source) and test (target) images.
For instance, the distribution of appearance for objects and scenes may vary in different cities, and even weather and lighting conditions can change significantly in the same city.
	In such cases, relying only on the supervised model that requires re-annotating per-pixel ground truths in different scenarios would entail prohibitively high labor cost.	

		\begin{figure}[t]
			\centering
			\includegraphics[width=1\linewidth]{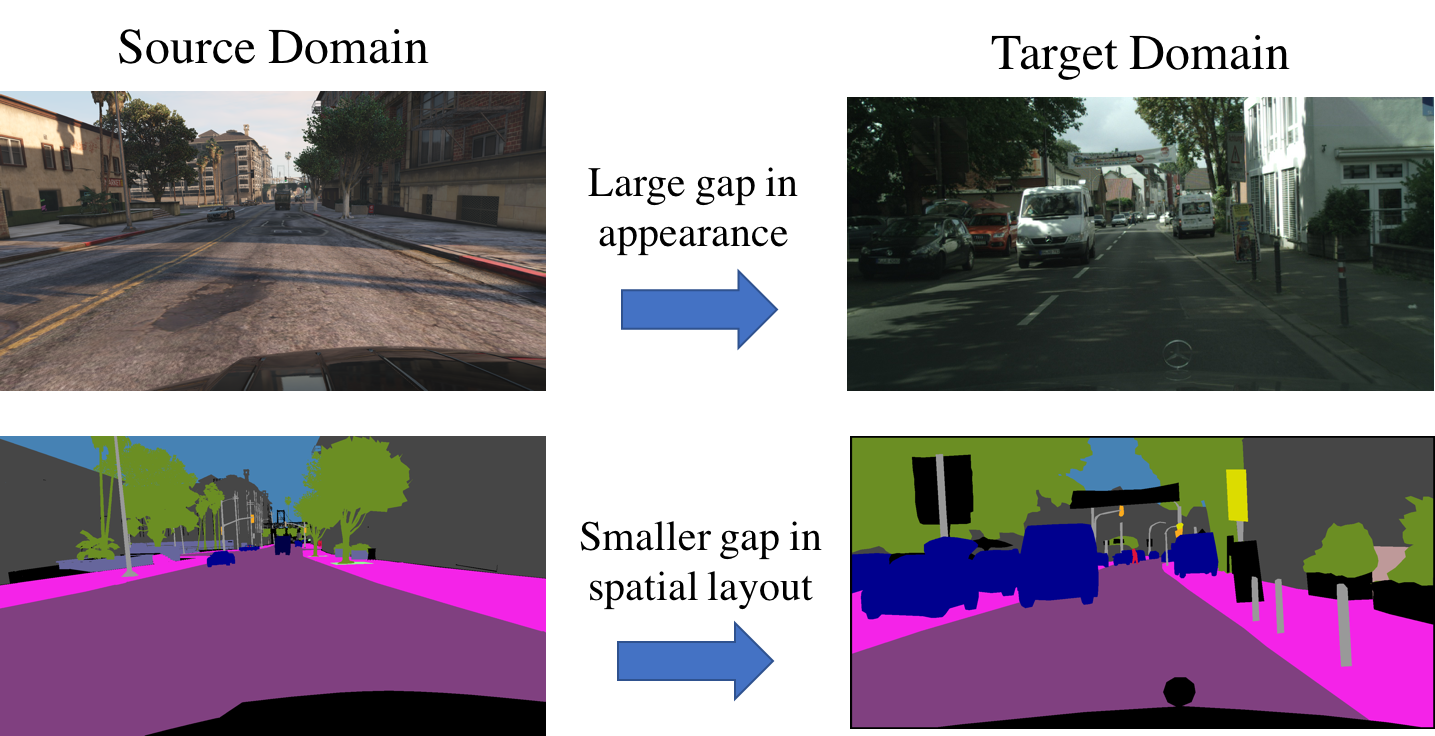}\\
			\caption{Our motivation of learning adaptation in the output space. While images may be very different in appearance, their outputs are structured and share many similarities, such as spatial layout and local context.}
			\label{figure: teaser}
			\vspace{-2mm}
		\end{figure}
	To address this issue, knowledge transfer or domain adaptation techniques have been proposed to close the gap between source and target domains, where annotations are not available in the target domain.
	For image classification, one effective approach is to align features across two domains \cite{ganin2016domain, long2015learning} such that the adapted features can generalize to both domains.
Similar efforts have been made for semantic segmentation via adversarial learning in the feature space~\cite{Chen_ICCV_2017, Hoffman_CoRR_2016}.
However, different from the image classification task, feature adaptation for semantic segmentation may suffer from the complexity of high-dimensional features that needs to encode diverse visual cues, including appearance, shape and context.
This motivates us to develop an effective method for adapting pixel-level prediction tasks rather than using feature adaptation.
In semantic segmentation, we note that the output space contains rich information, both spatially and locally.
For instance, even if images from two domains are very different in appearance, their segmentation outputs share a significant amount of similarities, e.g., spatial layout and local context (see Figure \ref{figure: teaser}).
Based on this observation, we address the pixel-level domain adaptation problem in the output (segmentation) space.

In this paper,  we propose an end-to-end CNN-based domain adaptation algorithm 
for semantic segmentation.
	%
Our formulation is based on adversarial learning in the output space, where the intuition is to 
directly make the predicted label distributions close to each other across source and target domains.
Based on the generative adversarial network (GAN)~\cite{Goodfellow_NIPS_2014, Radford_ICLR_2016, liu2016coupled}, the proposed model consists of 
two parts: 1) a segmentation model to predict output results, and 2) a discriminator to distinguish whether the input is from the source or target segmentation output.
With an adversarial loss, the proposed segmentation model aims to fool the discriminator, with the goal of generating similar distributions in the output space for either source or target images.

The proposed method also adapts features as the errors are back-propagated 
to the feature level from the output labels.
%
However, one concern is that lower-level features may not be adapted well as they are 
far away from the high-level output labels.
To address this issue, we develop a multi-level strategy by incorporating adversarial learning 
at different feature levels of the segmentation model.
For instance, we can use both \textit{conv5} and \textit{conv4} features to predict segmentation results in the output space.
Then two discriminators can be connected to each of the predicted output for multi-level adversarial learning.
We perform one-stage end-to-end  training for the segmentation model and discriminators jointly, without using any prior knowledge of the data in the target domain. 
In the testing phase, we can simply discard discriminators and use the adapted segmentation model on target images, with no extra computational requirements.
	
Due to the high labor cost of annotating segmentation ground truth, there has been great interest in large-scale synthetic datasets with annotations, e.g., GTA5~\cite{Richter_ECCV_2016} and SYNTHIA~\cite{Ros_CVPR_2016}.
As a result, one critical setting is to adapt the model trained on synthetic data to real-world datasets, such as Cityscapes~\cite{cityscapes}.
We follow this setting and conduct extensive experiments to validate the proposed domain adaptation method.
First, we use a strong baseline model that is able to generalize to different domains.
We note that a strong baseline facilitates real-world applications and can evaluate the limitation of 
the proposed adaptation approach.
Based on this baseline model, we show comparisons using adversarial adaptation in the feature and output spaces.
Furthermore, we show that the multi-level adversarial learning improves the results over single-level adaptation.
In addition to the synthetic-to-real setting, we show experimental results on the Cross-City dataset~\cite{Chen_ICCV_2017}, where annotations are provided in one city (source), while testing the model on another unseen city (target).
Overall, our method performs favorably against state-of-the-art algorithms on numerous benchmark datasets under different settings.
	
The contributions of this work are as follows.
First, we propose a domain adaptation method for pixel-level semantic segmentation via adversarial learning.
	Second, we demonstrate that adaptation in the output (segmentation) space can effectively align scene layout and local context between source and target images.
	Third, a multi-level adversarial learning scheme is developed to adapt features at different levels of the segmentation model, which leads to improved performance.
	
	
	%
	%
	%
	%
	%
	%

\section{Related Work}
\vspace{-1mm}	
{\flushleft {\bf Semantic Segmentation.}}
State-of-the-art semantic segmentation methods are mainly based on the 
recent advances of deep neural networks.
As proposed by Long~\etal\cite{Long_CVPR_2015}, one can transform a classification CNN (e.g., AlexNet~\cite{alexnet}, VGG~\cite{vgg}, or ResNet~\cite{He_CVPR_2016}) to a fully-convolutional network (FCN) for semantic segmentation.
Numerous methods have since been developed to improve this model by utilizing context information~\cite{hung2017scene, pspnet} or enlarging receptive fields~\cite{deeplab,Yu_ICLR_2016}.
To train these advanced networks, a substantial amount of dense pixel annotations must be collected in order to match the model capacity of deep CNNs.
	%
	%
As a result,  weakly and semi-supervised approaches~\cite{dai2015boxsup, hong2015decoupled, khoreva_CVPR17, papandreou2015weakly, pathak2015constrained} are proposed in recent years to reduce the heavy labeling cost of collecting segmentation ground truths.
	%
	%
However, in most real-world applications, it is difficult to obtain weak annotations and  
the trained model may not generalize well to unseen image domains.
	
Another approach to tackle the annotation problem is to construct synthetic datasets based on  rendering, e.g., GTA5~\cite{Richter_ECCV_2016} and SYNTHIA~\cite{Ros_CVPR_2016}.
While the data collection is less costly since the pixel-level annotation can be done with a partially automated process, 
these datasets are usually used in conjunction with real-world datasets for joint learning to improve the performance.
However, when training solely on the synthetic dataset, the model does not generalize well to real-world data, mainly due to the large domain shift between synthetic images and real-world images, i.e., appearance differences are still significant with current rendering techniques.
Although synthesizing more realistic images can decrease the domain shift, it is necessary to use domain adaptation to narrow the performance gap.
	\begin{figure*}[ht]
		\centering
		\includegraphics[width=0.8\linewidth]{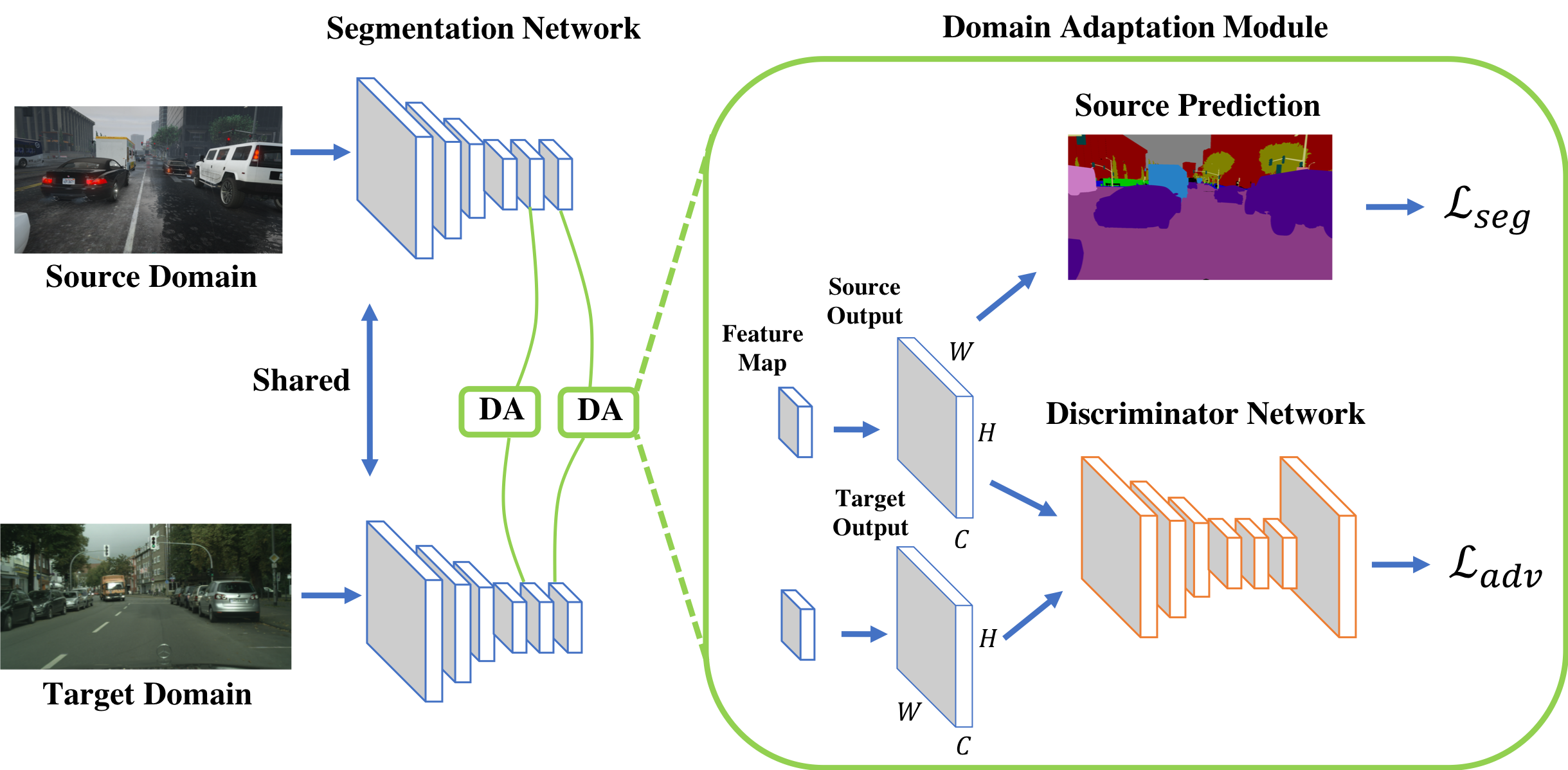}\\
		\vspace{1mm}
		\caption{
			Algorithmic overview.
			Given images with the size $H$ by $W$ in source and target domains, we pass them through the segmentation network to obtain output predictions.
			For source predictions with $C$ categories, a segmentation loss is computed based on the source ground truth.
			To make target predictions closer to the source ones, we utilize a discriminator to distinguish whether the input is from the source or target domain.
			Then an adversarial loss is calculated on the target prediction and is back-propagated to the segmentation network.
			We call this process as one adaptation module, and we illustrate our proposed multi-level adversarial learning by adopting two adaptation modules at two different levels here.
		}
		\label{figure: framework}
		\vspace{-2mm}
	\end{figure*}

\vspace{-2mm}	
	{\flushleft {\bf Domain Adaptation.}}
Domain adaptation methods for image classification have been developed to address the domain-shift problem between the source and target domains. 
Numerous methods~\cite{ganin2015unsupervised,ganin2016domain,long2015learning,long2016unsupervised,Sohn_ICCV_2017,tzeng2015simultaneous,tzeng2017adversarial} are 
developed based on CNN classifiers due to performance gain.
The main insight behind these approaches is to tackle the problem by aligning the feature distribution between source and target images.
Ganin~\etal~\cite{ganin2015unsupervised,ganin2016domain} 
propose the Domain-Adversarial Neural Network (DANN) to transfer the feature distribution.
	%
A number of variants have since been proposed with different loss functions~\cite{long2015learning,tzeng2015simultaneous, tzeng2017adversarial} or classifiers~\cite{long2016unsupervised}. 
Recently, the PixelDA method~\cite{Bousmalis_CVPR_2017} addresses domain adaptation for image classification by transferring the source images to target domain, thereby obtaining a simulated training set for target images.	
	%
	%
	
We note that domain adaptation for pixel-level prediction tasks have not been explored widely.
Hoffman~\etal~\cite{Hoffman_CoRR_2016} introduce the task of domain adaptation on semantic segmentation by applying adversarial learning (i.e., DANN) in a fully-convolutional way on feature representations and additional category constraints similar to the constrained CNN~\cite{pathak2015constrained}.
Other methods focus on adapting synthetic-to-real or cross-city images by adopting class-wise adversarial learning \cite{Chen_ICCV_2017} or label transfer \cite{Chen_ICCV_2017}.
Similar to the PixelDA method~\cite{Bousmalis_CVPR_2017}, one concurrent work, CyCADA~\cite{Hoffman_CoRR_2017} uses the CycleGAN~\cite{zhu2017unpaired} and transfers source domain images to the target domain with pixel alignment, thus
generating extra training data combined with feature space adversarial learning~\cite{Hoffman_CoRR_2016}.
	
Although feature space adaptation has been successfully applied to image classification, 
pixel-level tasks such as semantic segmentation remains challenging 
based on feature adaptation-based approaches.
In this paper, we use the property that pixel-level predictions are structured outputs that contain information spatially and locally,
	%
	%
	%
to propose an efficient domain adaptation algorithm through adversarial learning in the output space.
	
	%
	%

\section{Algorithmic Overview}
	
\subsection{Overview of the Proposed Model}
\label{sec:overview}
Our domain adaptation algorithm consists of two modules: a segmentation network $\mathbf{G}$ and the discriminator $\mathbf{D}_i$, where $i$ indicates the level of a discriminator in 
the multi-level adversarial learning.
Two sets of images $\in \mathbb{R}^{H \times W \times 3}$ from source and target domains are denoted
as $\{\mathcal{I}_S\}$ and $\{\mathcal{I}_T\}$.
We first forward the source image $I_s$ (with annotations) to the segmentation network for optimizing $\mathbf{G}$.
Then we predict the segmentation softmax output $P_t$ for the target image $I_t$ (without annotations).
	Since our goal is to make segmentation predictions $P$ of source and target images (i.e., $P_s$ and $P_t$) close to each other, we use these two predictions as the input to the discriminator $\mathbf{D}_i$ to distinguish whether the input is from the source or target domain.
With an adversarial loss on the target prediction, the network propagates gradients from $\mathbf{D}_i$ to $\mathbf{G}$, which would encourage $\mathbf{G}$ to generate similar segmentation distributions in the target domain to the source prediction.
Figure \ref{figure: framework} shows the overview of the proposed algorithm.

\subsection{Objective Function for Domain Adaptation}
	%

With the proposed network, 
we formulate the adaptation task containing two loss functions from both modules:
	\begin{equation}
	\mathcal{L}(I_s, I_t) = \mathcal{L}_{seg}(I_s) + \lambda_{adv}\mathcal{L}_{adv}(I_t),
	\label{eq:loss}
	\end{equation}
where $\mathcal{L}_{seg}$ is the cross-entropy loss using ground truth annotations in the source domain, and $\mathcal{L}_{adv}$ is the adversarial loss that adapts predicted segmentations of target images to the distribution of source predictions (see Section~\ref{sec:output-space}).
%
In~\eqref{eq:loss}, $\lambda_{adv}$ is the weight used to balance the two losses.
	%
	%
	%

	\section{Output Space Adaptation}
	\label{sec:output-space}
	%
	%
Different from image classification based on features \cite{ganin2016domain, long2015learning} that describe the global visual information of the image, high-dimensional features learned for semantic segmentation encodes complex representations.
As a result, adaptation in the feature space may not be the best choice for semantic segmentation.
	%
On the other hand, although segmentation outputs are in the low-dimensional space, they contain rich information, e.g., scene layout and context.
Our intuition is that no matter images are from the source or target domain, their segmentations should share strong similarities, spatially and locally.
Thus, we utilize this property to adapt low-dimensional softmax outputs of segmentation predictions via an adversarial learning scheme.
	\subsection{Single-level Adversarial Learning}
\vspace{-1mm}
	{\flushleft {\bf Discriminator Training.}}
Before introducing how to adapt the segmentation network via adversarial learning, we first describe the training objective for the discriminator.
Given the segmentation softmax output $P = \mathbf{G}(I) \in \mathbb{R}^{H \times W \times C}$, where $C$ is the number of categories, we forward $P$ to a fully-convolutional discriminator $\mathbf{D}$ using a cross-entropy loss $\mathcal{L}_{d}$ for the two classes (i.e., source and target).
	The loss can be written as:
	\begin{align}
	\label{eq:loss_d}
	\mathcal{L}_{d}(P) = - \sum_{h,w}{} (1-z) \log(\mathbf{D}(P)^{(h,w,0)}) \\ \notag
	+ z \log(\mathbf{D}(P)^{(h,w,1)}),
	\end{align}
	where $z = 0$ if the sample is drawn from the target domain, and $z = 1$ for the sample from the source domain.
	
\vspace{-3mm}
	{\flushleft {\bf Segmentation Network Training.}}
	First, we define the segmentation loss in \eqref{eq:loss} as the cross-entropy loss for images from the source domain:
	\begin{equation}
	\mathcal{L}_{seg}(I_s) = - \sum_{h,w}{} \sum_{c \in C}Y_s^{(h,w,c)} \log(P_s^{(h,w,c)}),
	\label{eq:loss_seg}
	\end{equation}
	where $Y_s$ is the ground truth annotations for source images and $P_s = \mathbf{G}(I_s)$ is the segmentation output.

Second, for images in the target domain, we forward them to $\mathbf{G}$ and obtain the prediction $P_t = \mathbf{G}(I_t)$.
To make the distribution of $P_t$ closer to $P_s$, we use an adversarial loss $\mathcal{L}_{adv}$ in \eqref{eq:loss} as:
	\begin{equation}
	\mathcal{L}_{adv}(I_t) = - \sum_{h,w}{} \log(\mathbf{D}(P_t)^{(h,w,1)}).
	\label{eq:loss_adv}
	\end{equation}
This loss is designed to train the segmentation network and fool the discriminator by maximizing the probability of the target prediction being considered as the source prediction.
	
	%
	
	\subsection{Multi-level Adversarial Learning}
	Although performing adversarial learning in the output space directly adapts predictions, low-level features may not be adapted well as they are far away from the output.
	%
Similar to the deep supervision method~\cite{Lee_AISTATS_2015} that uses auxiliary loss for semantic segmentation~\cite{pspnet}, we incorporate additional adversarial module in the low-level feature space to enhance the adaptation.
	The training objective for the segmentation network can be extended from \eqref{eq:loss} as:
	\begin{equation}
	\mathcal{L}(I_s, I_t) = \sum_i \lambda^i_{seg}\mathcal{L}^i_{seg}(I_s) + \sum_i \lambda^i_{adv}\mathcal{L}^i_{adv}(I_t),
	\label{eq:loss_multi}
	\end{equation}
where $i$ indicates the level used for predicting the segmentation output.
We note that, the segmentation output is still predicted in each feature space, before passing through individual discriminators for adversarial learning.
	Hence, $\mathcal{L}^i_{seg}(I_s)$ and $\mathcal{L}^i_{adv}(I_t)$ remain in the same form as in \eqref{eq:loss_seg} and \eqref{eq:loss_adv}, respectively.
	%
	%
	Based on \eqref{eq:loss_multi}, we optimize the following min-max criterion:
	\begin{equation}
	\max_{\mathbf{D}} \min_{\mathbf{G}} \mathcal{L}(I_s, I_t).
	\label{eq:optimize}
	\end{equation}
	The ultimate goal is to minimize the segmentation loss in $\mathbf{G}$ for source images, while maximizing the probability of target predictions being considered as source predictions.
	%
	%
	\begin{table*} [t]
		\caption{Results of adapting GTA5 to Cityscapes. We first compare our results using single-level adversarial learning in the output space with other state-of-the-art algorithms with the VGG-16 based model.
			Then we adopt the ResNet-101 based model and present ablation study on different components of our proposed method.	
	}
		\vspace{1mm}
		\label{table:gta5_all}
		\small
		\centering
		\renewcommand{\arraystretch}{1.1}
		\setlength{\tabcolsep}{2.2pt}
		\begin{tabular}{lcccccccccccccccccccc}
			\toprule
			
			& \multicolumn{20}{c}{GTA5 $\rightarrow$ Cityscapes} \\
			\midrule
			
			Method & \rotatebox{90}{road} & \rotatebox{90}{sidewalk} & \rotatebox{90}{building} & \rotatebox{90}{wall} & \rotatebox{90}{fence} & \rotatebox{90}{pole} & \rotatebox{90}{light} & \rotatebox{90}{sign} & \rotatebox{90}{veg} & \rotatebox{90}{terrain} & \rotatebox{90}{sky} & \rotatebox{90}{person} & \rotatebox{90}{rider} & \rotatebox{90}{car} & \rotatebox{90}{truck} & \rotatebox{90}{bus} & \rotatebox{90}{train} & \rotatebox{90}{mbike} & \rotatebox{90}{bike} & mIoU\\
			
			\midrule
			
			FCNs in the Wild \cite{Hoffman_CoRR_2016} & 70.4 & 32.4 & 62.1 & 14.9 & 5.4 & 10.9 & 14.2 & 2.7 & 79.2 & 21.3 & 64.6 & 44.1 & 4.2 & 70.4 & 8.0 & 7.3 & 0.0 & 3.5 & 0.0 & 27.1 \\
			
			CDA \cite{Zhang_ICCV_2017} & 74.9 & 22.0 & 71.7 & 6.0 & 11.9 & 8.4 & 16.3 & 11.1 & 75.7 & 13.3 & 66.5 & 38.0 & \textbf{9.3} & 55.2 & 18.8 & 18.9 & 0.0 & \textbf{16.8} & \textbf{14.6} & 28.9 \\
			
			CyCADA (feature) \cite{Hoffman_CoRR_2017}  & 85.6 & 30.7 & 74.7 & 14.4 & 13.0 & 17.6 & 13.7 & 5.8 & 74.6 & 15.8 & 69.9 & 38.2 & 3.5 & 72.3 & 16.0 & 5.0 & 0.1 & 3.6 & 0.0 & 29.2 \\
			
			CyCADA (pixel) \cite{Hoffman_CoRR_2017} & 83.5 & \textbf{38.3} & 76.4 & 20.6 & 16.5 & 22.2 & \textbf{26.2} & \textbf{21.9} & \textbf{80.4} & 28.7 & 65.7 & \textbf{49.4} & 4.2 & 74.6 & 16.0 & 26.6 & \textbf{2.0} & 8.0 & 0.0 & 34.8 \\ 
			
			
			Ours (singel-level) &  \textbf{87.3} & 29.8 & \textbf{78.6} & \textbf{21.1} & \textbf{18.2} & \textbf{22.5} & 21.5 & 11.0 & 79.7 & \textbf{29.6} & \textbf{71.3} & 46.8 & 6.5 & \textbf{80.1} & \textbf{23.0} & \textbf{26.9} & 0.0 & 10.6 & 0.3 & \textbf{35.0} \\ 
			
			\midrule
			Baseline (ResNet) & 75.8 & 16.8 & 77.2 & 12.5 & 21.0 & 25.5 & 30.1 & 20.1 & 81.3 & 24.6 & 70.3 & 53.8 & 26.4 & 49.9 & 17.2 & 25.9 & 6.5 & 25.3 & \textbf{36.0} & 36.6 \\
			
			Ours (feature) & 83.7 & 27.6 & 75.5 & 20.3 & 19.9 & \textbf{27.4} & 28.3 & \textbf{27.4} & 79.0 & 28.4 & 70.1 & 55.1 & 20.2 & 72.9 & 22.5 & \textbf{35.7} & \textbf{8.3} & 20.6 & 23.0 & 39.3 \\
			
			Ours (single-level) & \textbf{86.5} & 25.9 & 79.8 & 22.1 & 20.0 & 23.6 & 33.1 & 21.8 & 81.8 & 25.9 & \textbf{75.9} & 57.3 & 26.2 & \textbf{76.3} & 29.8 & 32.1 & 7.2 & 29.5 & 32.5 & 41.4 \\
			
			Ours (multi-level) & \textbf{86.5} & \textbf{36.0} & \textbf{79.9} & \textbf{23.4} & \textbf{23.3} & 23.9 & \textbf{35.2} & 14.8 & \textbf{83.4} & \textbf{33.3} & 75.6 & \textbf{58.5} & \textbf{27.6} & 73.7 & \textbf{32.5} & 35.4 & 3.9 & \textbf{30.1} & 28.1 & \textbf{42.4} \\
			
			%
			
			\bottomrule
		\end{tabular}
	\vspace{-1mm}
	\end{table*}

\section{Network Architecture and Training}
\label{sec:network}
\vspace{-1mm}
{\flushleft {\bf Discriminator.}}
For the discriminator, we use an architecture similar to \cite{Radford_ICLR_2016}  
but utilize all fully-convolutional layers to retain the spatial information.
The network consists of 5 convolution layers with kernel $4 \times 4$ and stride of 2, where the channel number is \{64, 128, 256, 512, 1\}, respectively.
	Except for the last layer, each convolution layer is followed by a leaky ReLU~\cite{maas2013rectifier} parameterized by $0.2$.
	%
	%
We do not use any batch-normalization layers~\cite{Ioffe_ICML_2015} as we jointly train the discriminator with the segmentation network using a small batch size.

\vspace{-3mm}
	{\flushleft {\bf Segmentation Network.}}
It is essential to build upon a good baseline model to achieve high-quality segmentation results \cite{deeplab, Yu_ICLR_2016, pspnet}.
We adopt the DeepLab-v2~\cite{deeplab} framework with ResNet-101~\cite{He_CVPR_2016} model pre-trained on ImageNet~\cite{imagenet} as our segmentation baseline network.
However, we do not use the multi-scale fusion strategy~\cite{deeplab} due to the memory issue.
Similar to the recent work on semantic segmentation~\cite{deeplab, Yu_ICLR_2016}, we remove the last classification layer and modify the stride of the last two convolution layers from 2 to 1, making the resolution of the output feature maps effectively $1/8$ times the input image size.
To enlarge the receptive field, we apply dilated convolution layers~\cite{Yu_ICLR_2016} in \textit{conv4} and \textit{conv5} layers with a stride of 2 and 4, respectively.
	After the last layer, we use the Atrous Spatial Pyramid Pooling (ASPP)~\cite{deeplab} as the final classifier.
Finally, we apply an up-sampling layer along with the softmax output to match the size of the input image.
Based on this architecture, our segmentation model achieves 65.1\% mean intersection-over-union (IoU) when trained on the Cityscapes~\cite{cityscapes} training set and tested on the Cityscapes validation set.

\vspace{-2mm}
	{\flushleft {\bf Multi-level Adaptation Model.}}
We construct the above-mentioned discriminator and segmentation network as our single-level adaptation model.
For the multi-level structure, we extract feature maps from the \textit{conv4} layer and add an ASPP module as the auxiliary classifier.
	Similarly, a discriminator with the same architecture is added for adversarial learning.
Figure \ref{figure: framework} shows the proposed multi-level adaptation model.
In this paper, we use two levels due to the balance of its efficiency and accuracy.
	%
	%
	%
\vspace{-2mm}
	{\flushleft {\bf Network Training.}}
	%
To train the proposed single/multi-level adaptation model, we find that jointly training the segmentation network and discriminators in one stage is effective.
	In each training batch, we first forward the source image $I_s$ to optimize the segmentation network for $\mathcal{L}_{seg}$ in \eqref{eq:loss_seg} and generate the output $P_s$.
	For the target image $I_t$, we obtain the segmentation output $P_t$, and pass it along with $P_s$ to the discriminator for optimizing $\mathcal{L}_{d}$ in \eqref{eq:loss_d}.
In addition, we compute the adversarial loss $\mathcal{L}_{adv}$ in \eqref{eq:loss_adv} for the target prediction $P_t$.
For the multi-level training objective in \eqref{eq:loss_multi}, we simply repeat the same procedure for each adaptation module.

We implement our network using the PyTorch toolbox  
on a single Titan X GPU with 12 GB memory.
To train the segmentation network, we use the Stochastic Gradient Descent (SGD) optimizer with Nesterov acceleration where the momentum is 0.9 and the weight decay is $5 \times 10^{-4}$.
The initial learning rate is set as $2.5 \times 10^{-4}$ and is decreased using the polynomial decay with power of 0.9 as mentioned in \cite{deeplab}.
For training the discriminator, we use the Adam optimizer~\cite{Kingma_ICLR_2015} with the learning rate as $10^{-4}$ and the same polynomial decay as the segmentation network. The momentum is set as 0.9 and 0.99.
%

	%
	
%
\begin{table} [t]
	\caption{
		Performance gap between the adapted model and the fully-supervised (oracle) model.
		We first compare results with state-of-the-art methods using the VGG based model, and then show our result using the ResNet one.
	}
	\vspace{1mm}
	\label{table:gta5_gap}
	\small
	\centering
	\renewcommand{\arraystretch}{1.0}
	\setlength{\tabcolsep}{2.5pt}
	\begin{tabular}{lcccc}
		\toprule
		
		& \multicolumn{3}{c}{GTA5 $\rightarrow$ Cityscapes} \\
		\midrule
		
		method & Baseline & Adapt & Oracle & mIoU Gap \\
		
		\midrule
		
		FCNs in the Wild \cite{Hoffman_CoRR_2016} & \multirow{5}{*}{VGG-16} & 27.1 & 64.6 & -37.5 \\
		
		CDA \cite{Zhang_ICCV_2017}  &  & 28.9 & 60.3 & -31.4 \\
		
		CyCADA (feature) \cite{Hoffman_CoRR_2017} &  & 29.2 & 60.3 & -30.5 \\
		
		CyCADA (pixel) \cite{Hoffman_CoRR_2017}  &  & 34.8 & 60.3 & -24.9 \\
		
		Ours (single-level) &  & 35.0 & 61.8 & -25.2 \\
		
		\midrule
		
		Ours (multi-level)& ResNet-101 & 42.4 & 65.1 & -22.7 \\
		\bottomrule
	\end{tabular}
	\vspace{-3mm}
\end{table}
\section{Experimental Results}
In this section, we present experimental results to validate the proposed domain adaptation method for semantic segmentation under different settings.
First, we show evaluations of the model trained on synthetic datasets (i.e., GTA5~\cite{Richter_ECCV_2016} and SYNTHIA~\cite{Ros_CVPR_2016}) 
and test the adapted model on real-world images from the Cityscapes~\cite{cityscapes} dataset.
Extensive experiments including comparisons to the state-of-the-art methods and ablation study are also conducted, e.g., adaptation in the feature/output spaces and single/multi-level adversarial learning.
	%
Second, we carry out experiments on the Cross-City dataset~\cite{Chen_ICCV_2017}, where the model is trained on one city and adapted to another city without using annotations.
In all the experiments, the IoU metric is used.
	%
The code and model are available at \url{https://github.com/wasidennis/AdaptSegNet}.
	\subsection{GTA5}
The GTA5 dataset~\cite{Richter_ECCV_2016} consists of $24966$ images with the resolution of $1914 \times 1052$ synthesized from the video game based on the city of Los Angeles.
	The ground truth annotations are compatible with the Cityscapes dataset~\cite{cityscapes} that contains 19 categories.
	Following~\cite{Hoffman_CoRR_2016}, we use the full set of GTA5 and adapt the model to the Cityscapes training set with 2975 images.
	During testing, we evaluate on the Cityscapes validation set with 500 images.
	%

\vspace{-3mm}
	{\flushleft {\bf Overall Results.}}
	We present adaptation results in Table \ref{table:gta5_all} with comparisons to the state-of-the-art domain adaptation methods \cite{Hoffman_CoRR_2017, Hoffman_CoRR_2016, Zhang_ICCV_2017}.
For these approaches, the baseline model is trained using VGG-based architectures \cite{Long_CVPR_2015, Yu_ICLR_2016}.
To fairly evaluate our method, we first use the same baseline architecture (VGG-16) and train our model with the proposed single-level adaptation module.
Table \ref{table:gta5_all} shows that our method performs favorably against the other algorithms. 
While these methods all have feature adaptation modules, our results
show that adapting the model in the output space achieves better performance.
We note that CyCADA \cite{Hoffman_CoRR_2017} has a pixel adaptation module by transforming source domain images to the target domain and hence obtains additional training samples.
Although this strategy achieves a similar performance as ours, one can always apply pixel transformation combined with our output space adaptation to improve the results.
%
%

\begin{figure*}[t]
		\centering
		\begin{tabular}
			{@{\hspace{0mm}}c@{\hspace{1mm}} @{\hspace{0mm}}c@{\hspace{1mm}}
				@{\hspace{0mm}}c@{\hspace{1mm}} @{\hspace{0mm}}c@{\hspace{1mm}}
				@{\hspace{0mm}}c@{\hspace{0mm}}
			}
			
			\includegraphics[width=0.19\linewidth]{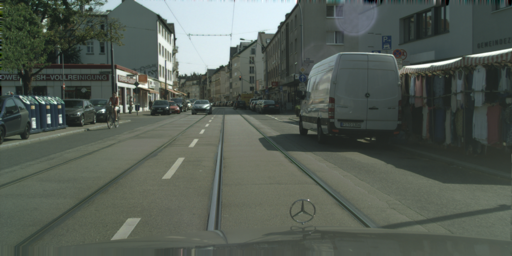} &
			\includegraphics[width=0.19\linewidth]{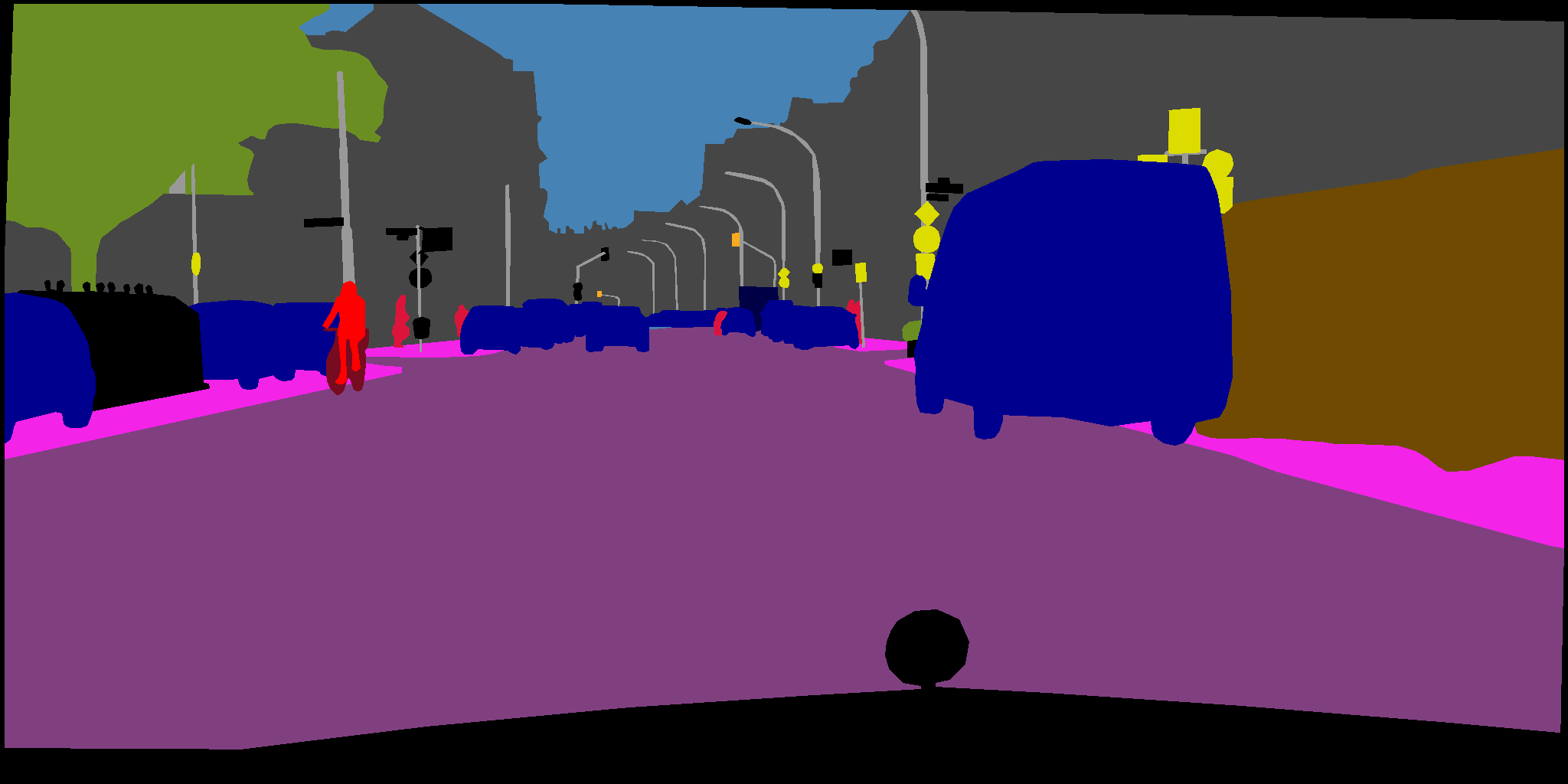} &
			\includegraphics[width=0.19\linewidth]{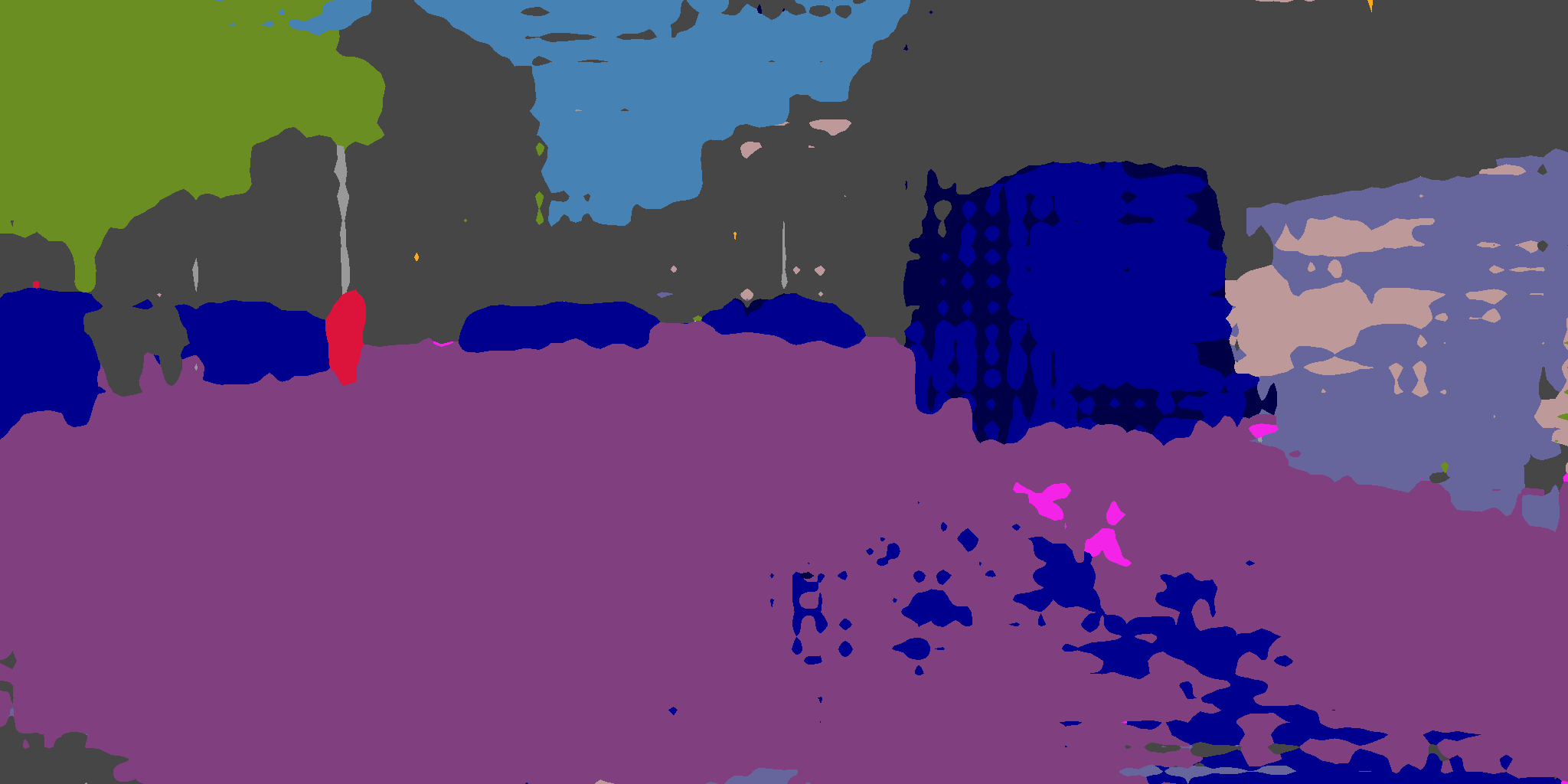} &
			\includegraphics[width=0.19\linewidth]{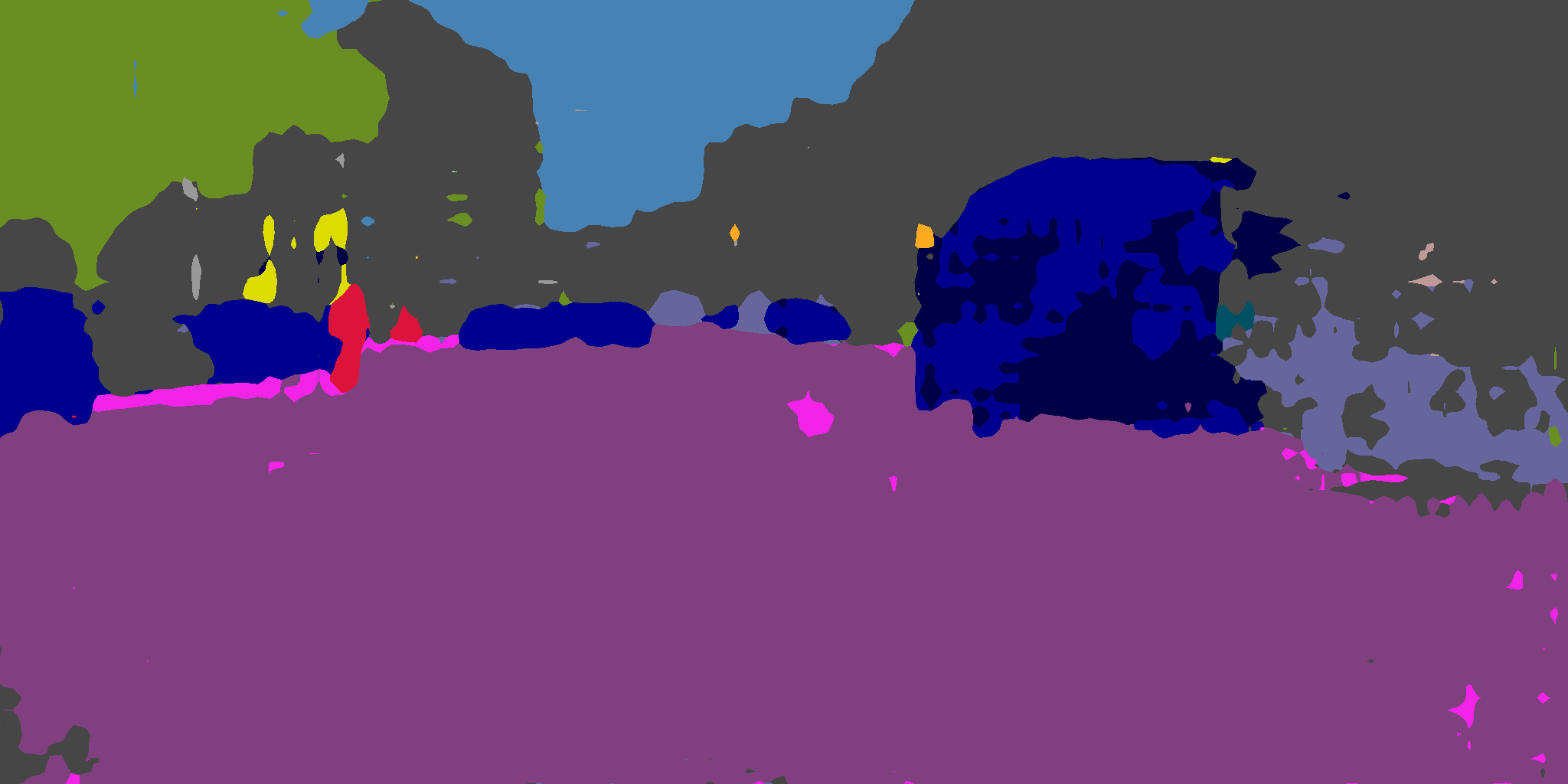} &
			\includegraphics[width=0.19\linewidth]{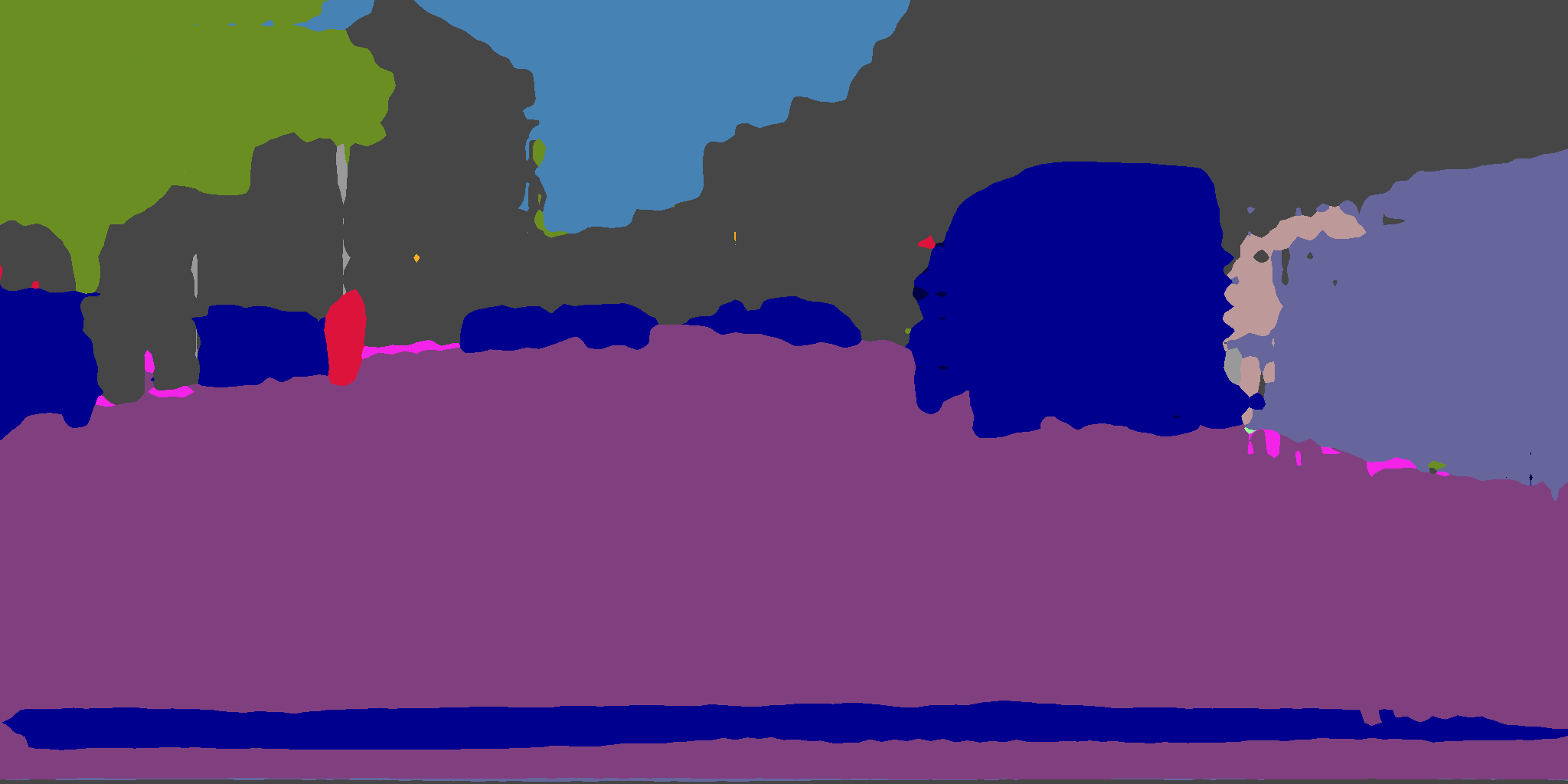} \\		
			
			\includegraphics[width=0.19\linewidth]{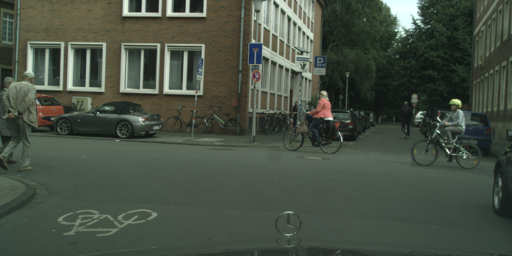} &
			\includegraphics[width=0.19\linewidth]{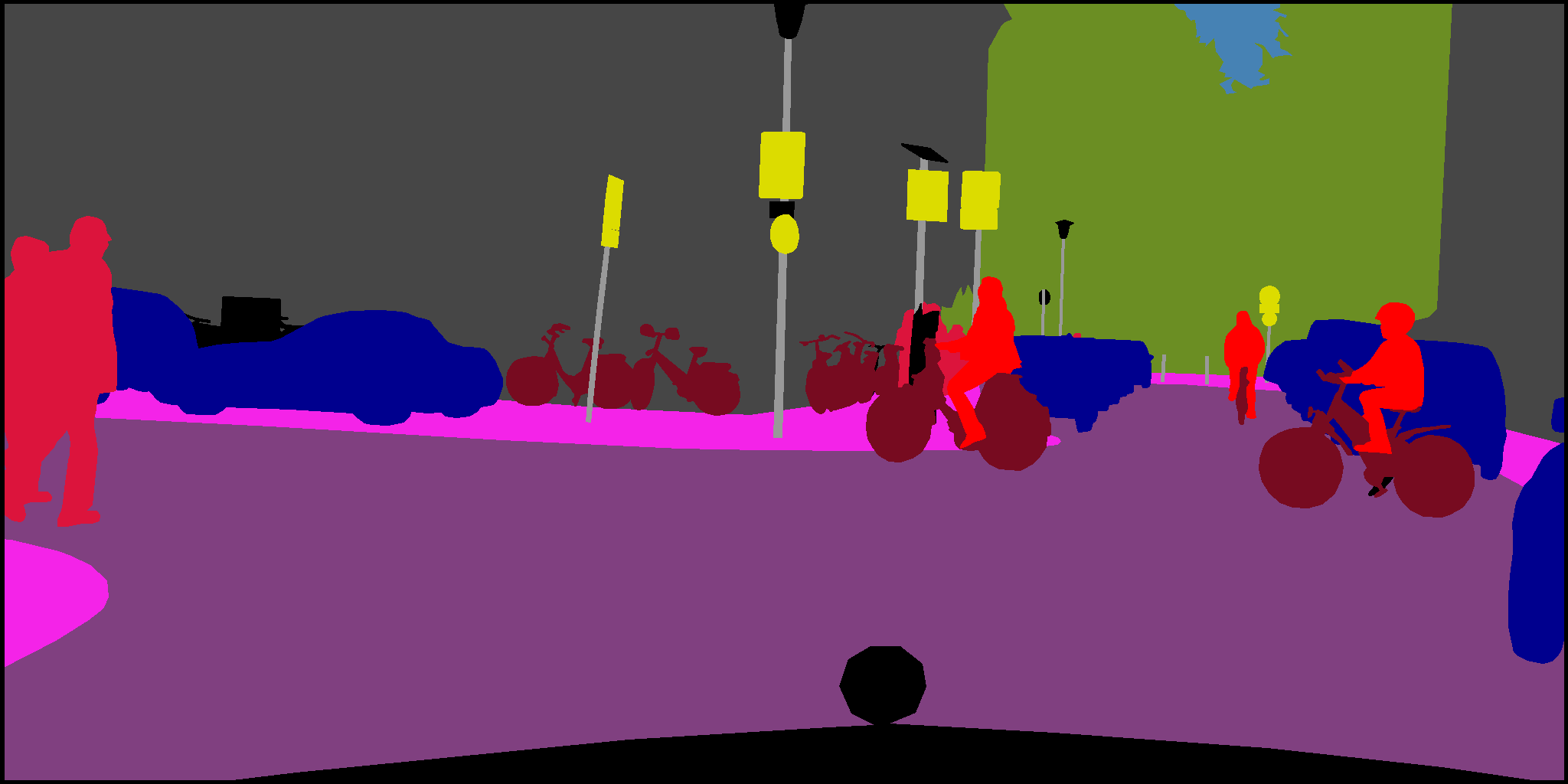} &
			\includegraphics[width=0.19\linewidth]{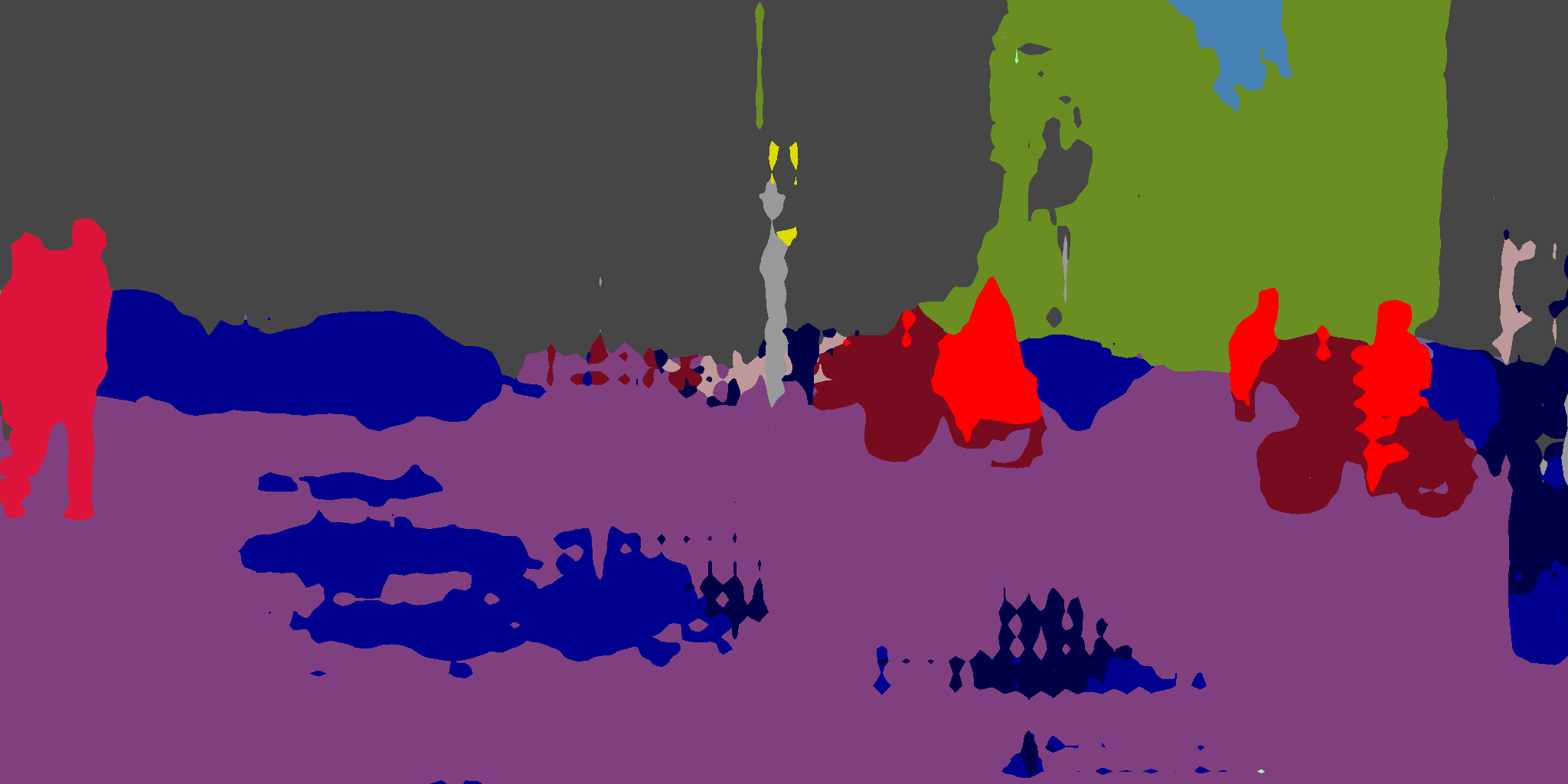} &
			\includegraphics[width=0.19\linewidth]{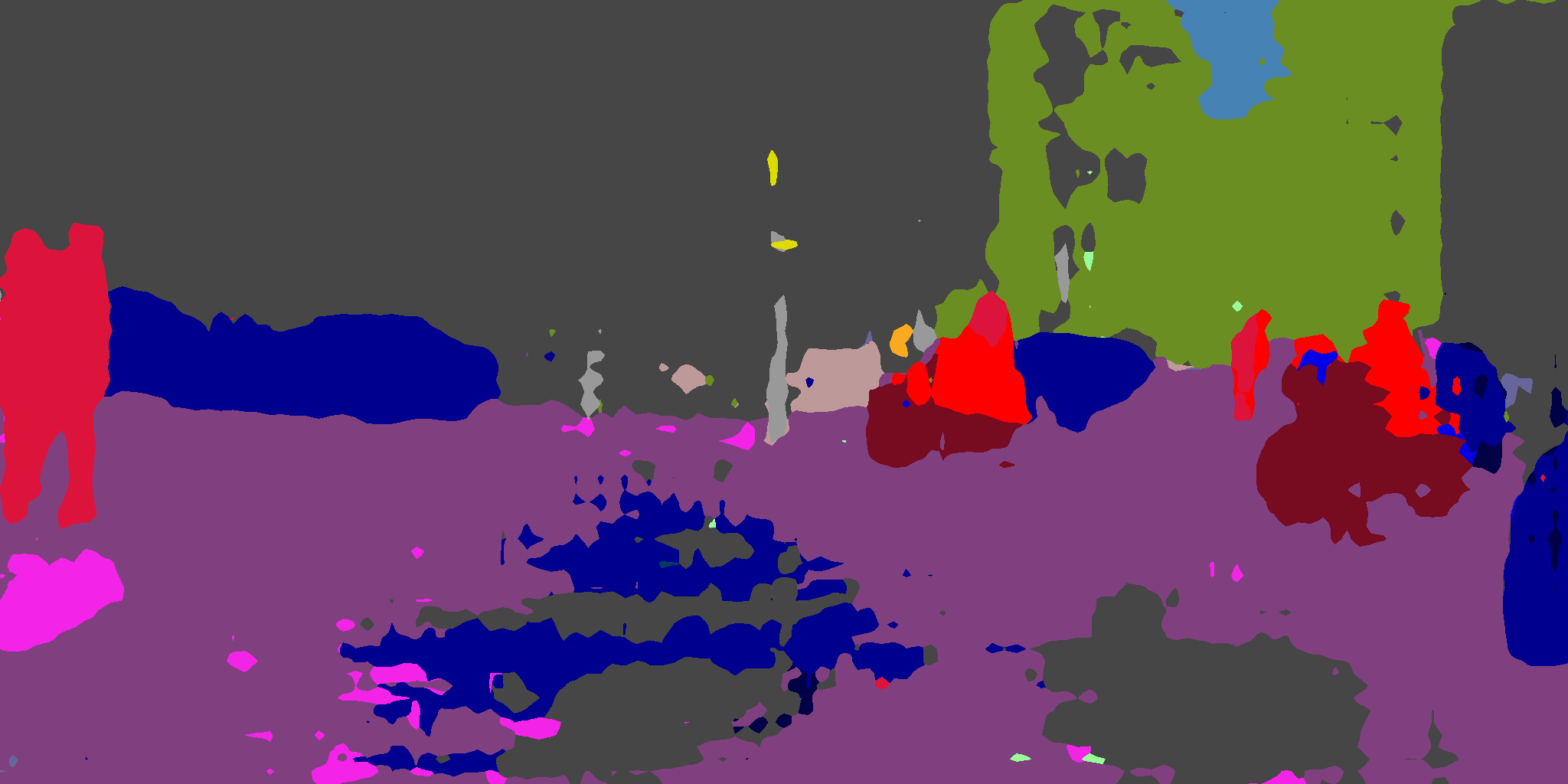} &
			\includegraphics[width=0.19\linewidth]{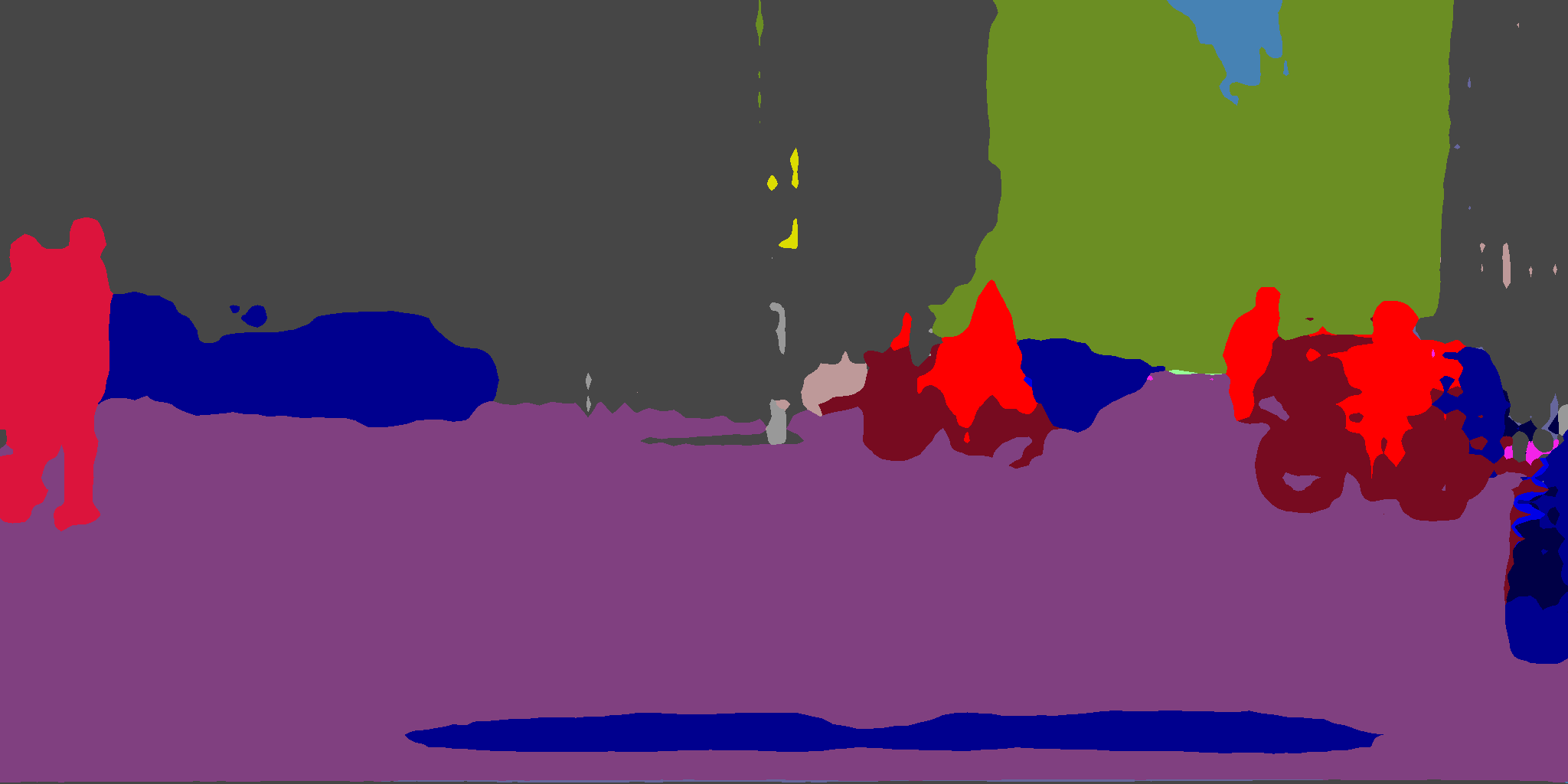} \\
			
			\includegraphics[width=0.19\linewidth]{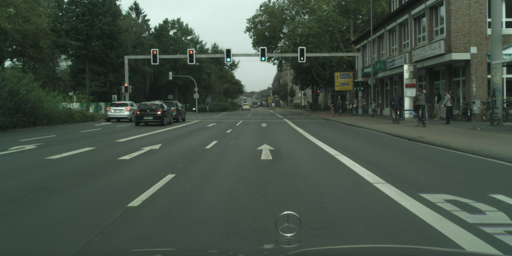} &
			\includegraphics[width=0.19\linewidth]{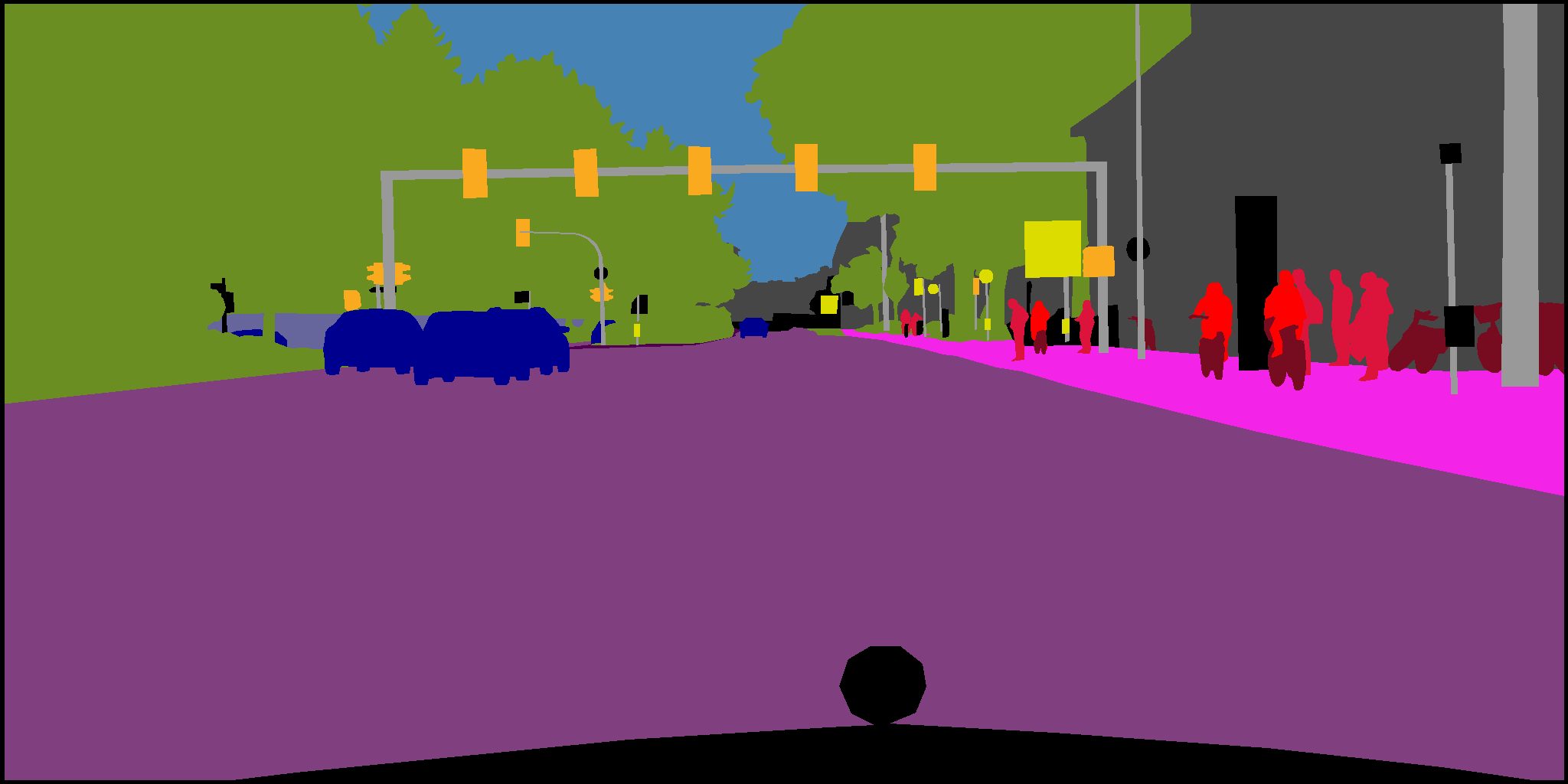} &
			\includegraphics[width=0.19\linewidth]{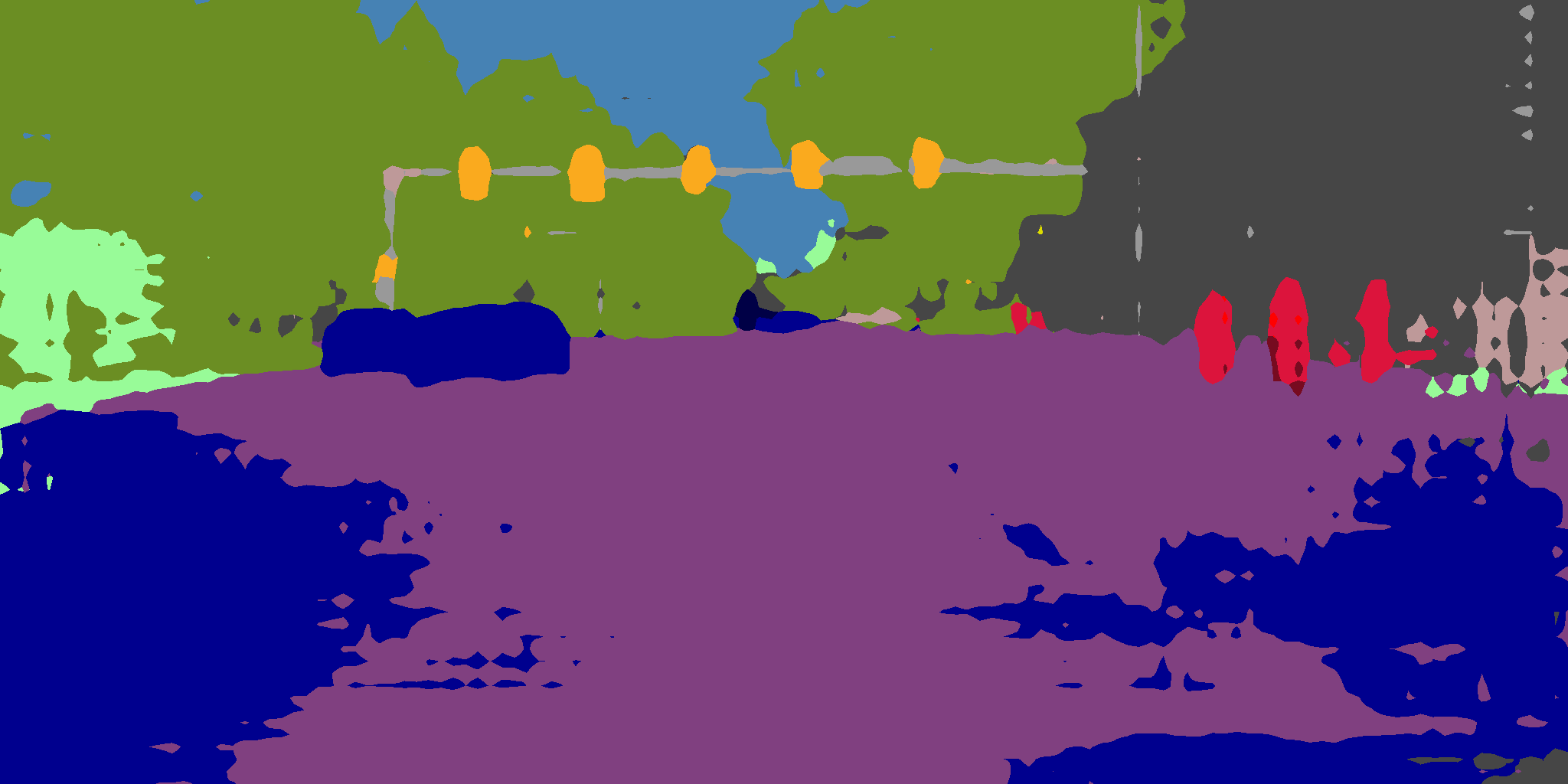} &
			\includegraphics[width=0.19\linewidth]{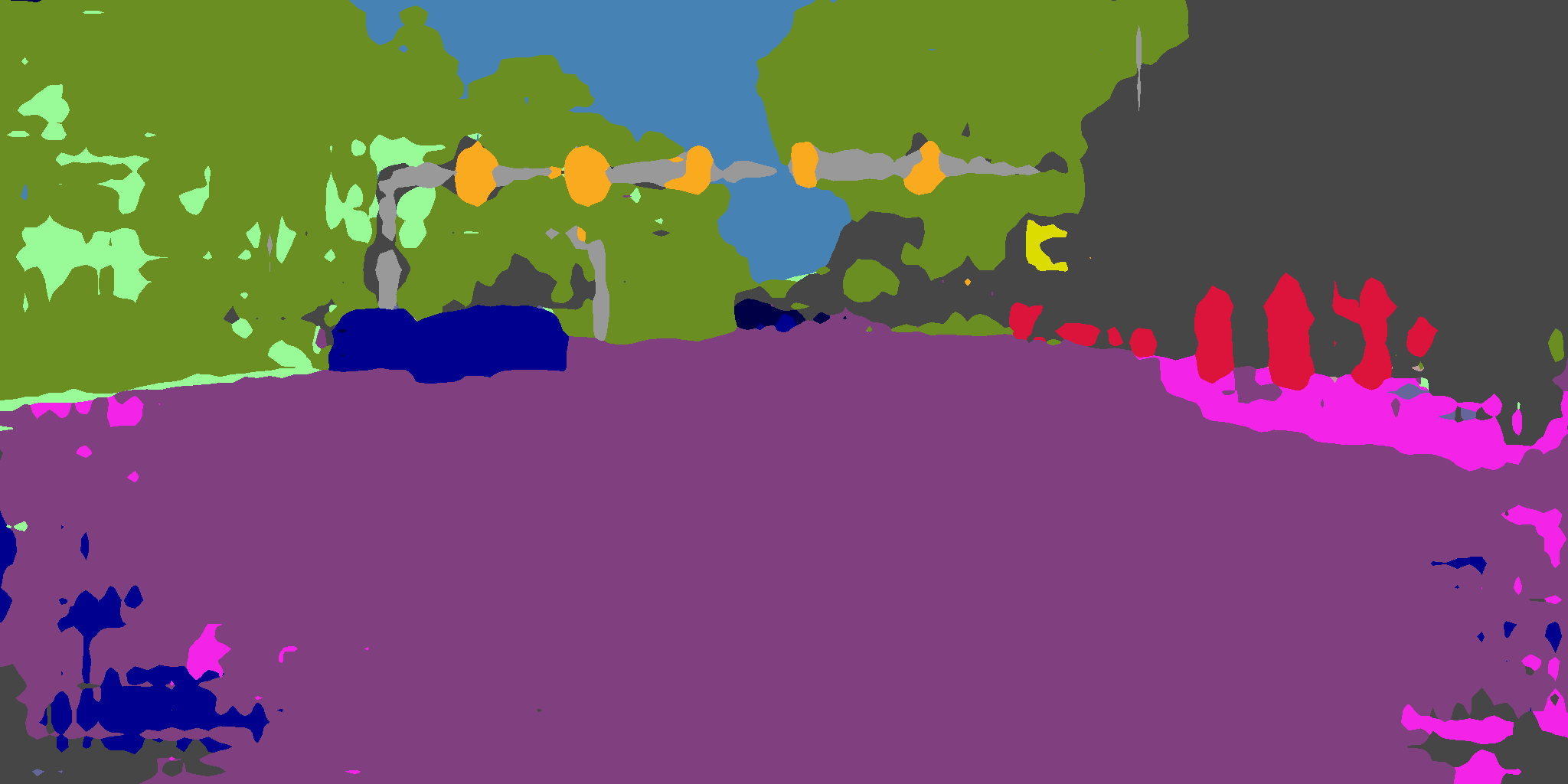} &
			\includegraphics[width=0.19\linewidth]{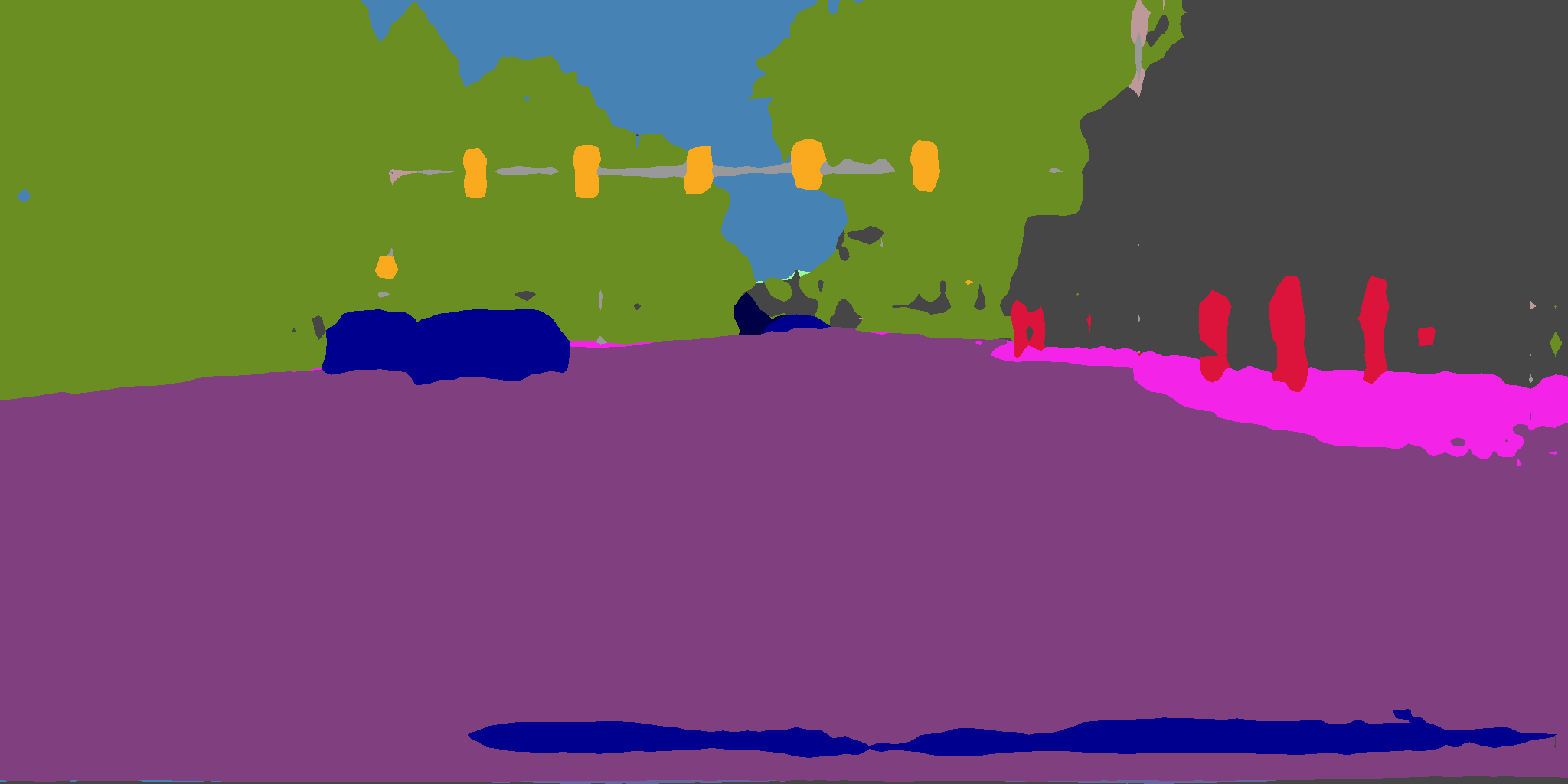} \\
			
			\includegraphics[width=0.19\linewidth]{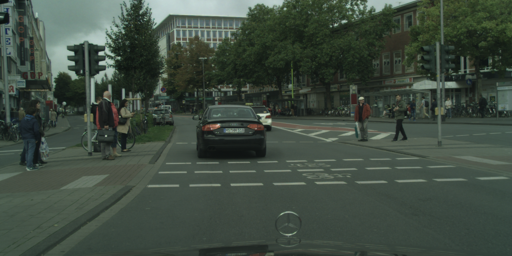} &
			\includegraphics[width=0.19\linewidth]{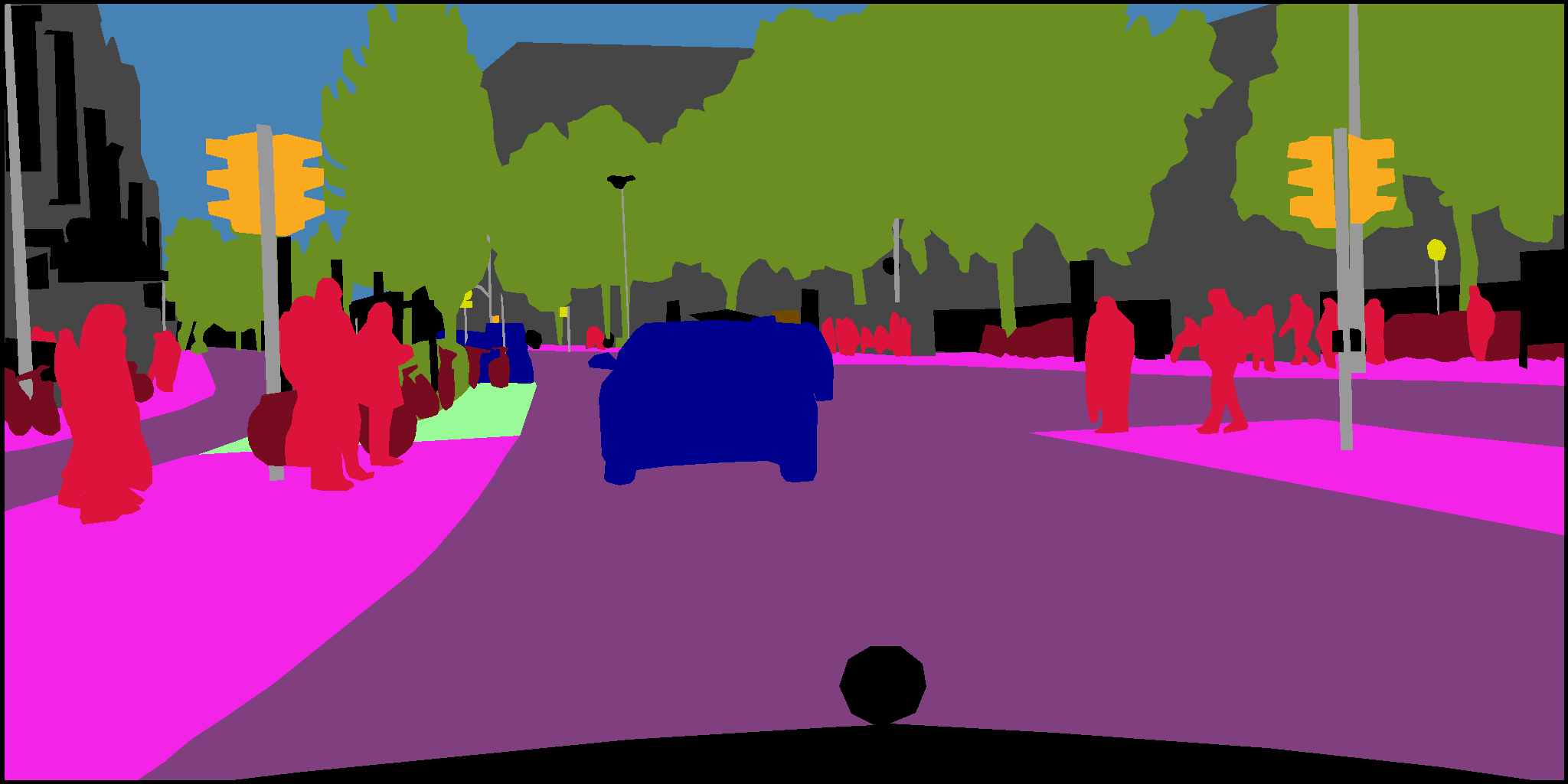} &
			\includegraphics[width=0.19\linewidth]{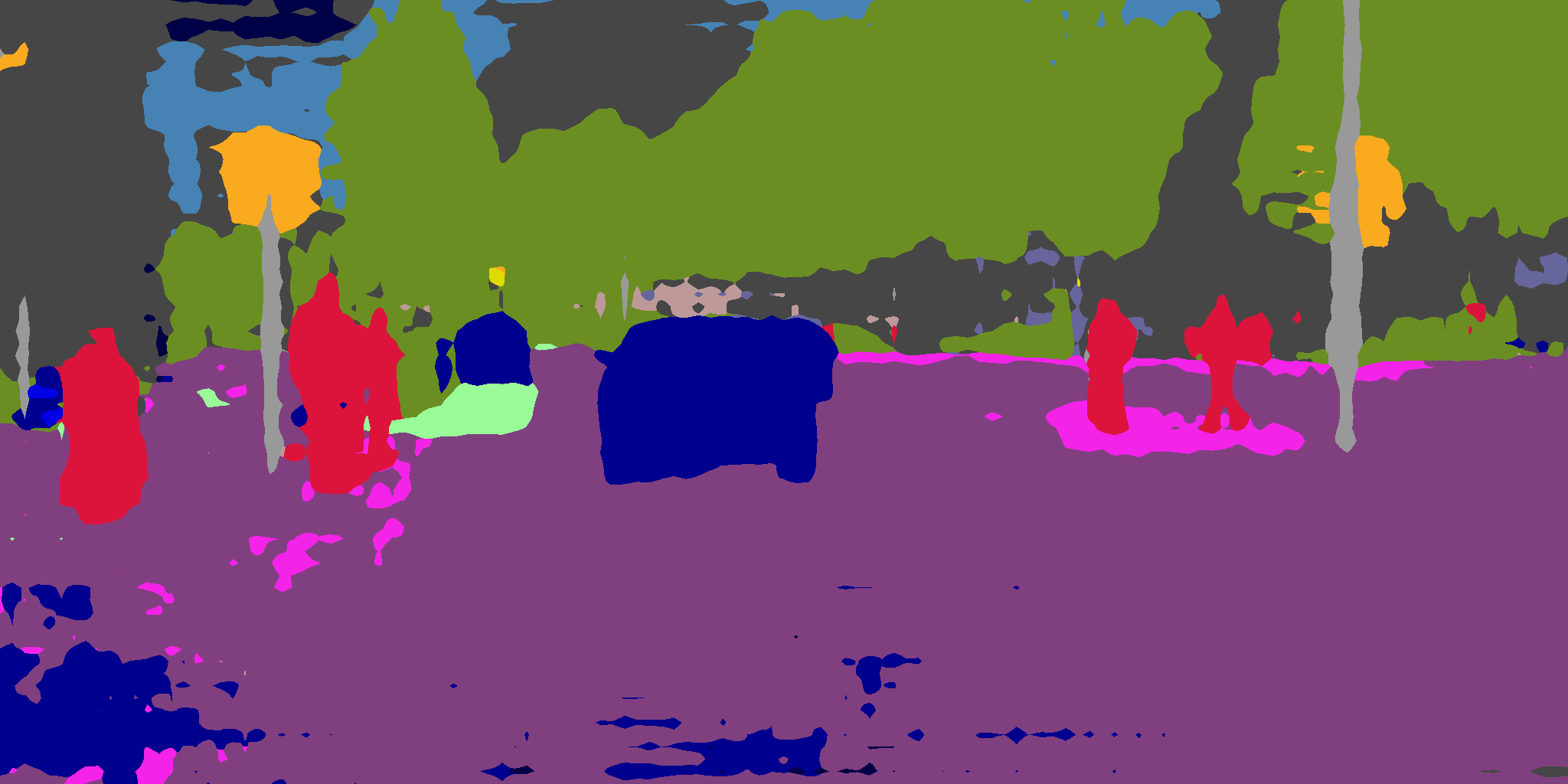} &
			\includegraphics[width=0.19\linewidth]{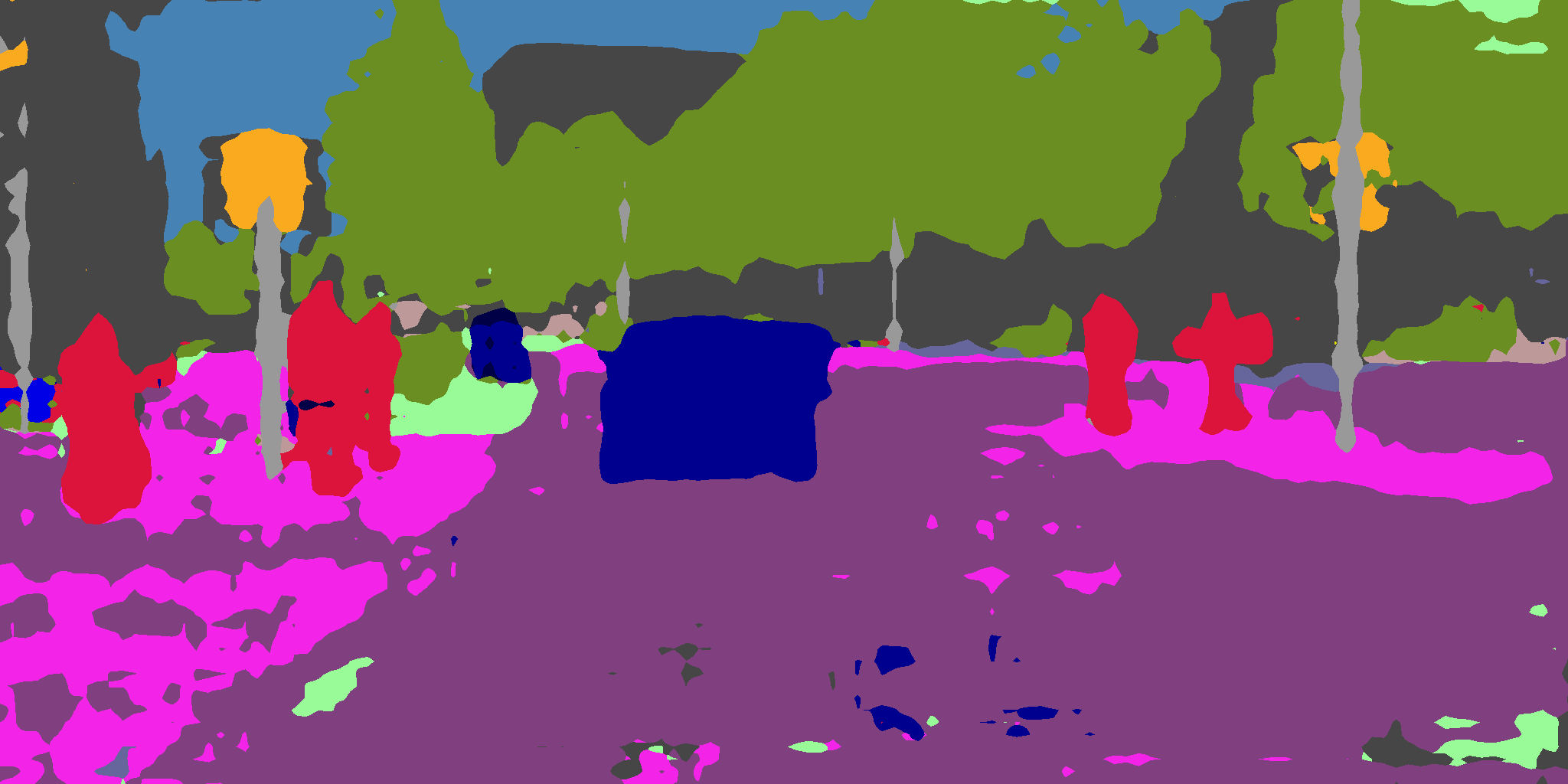} &
			\includegraphics[width=0.19\linewidth]{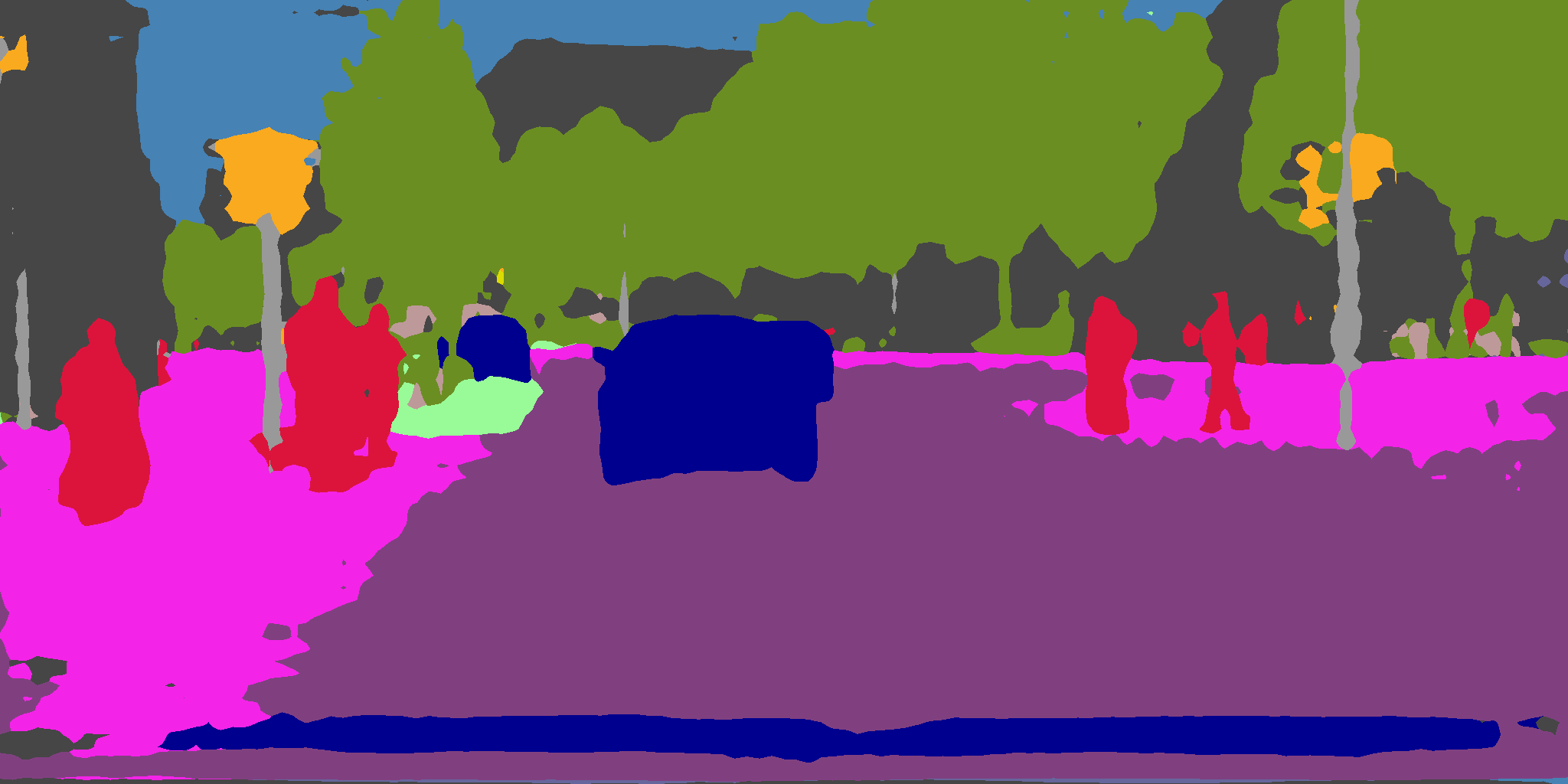} \\
			
			\includegraphics[width=0.19\linewidth]{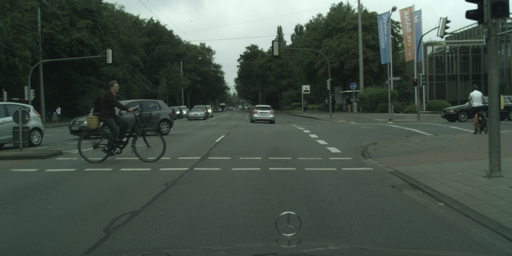} &
			\includegraphics[width=0.19\linewidth]{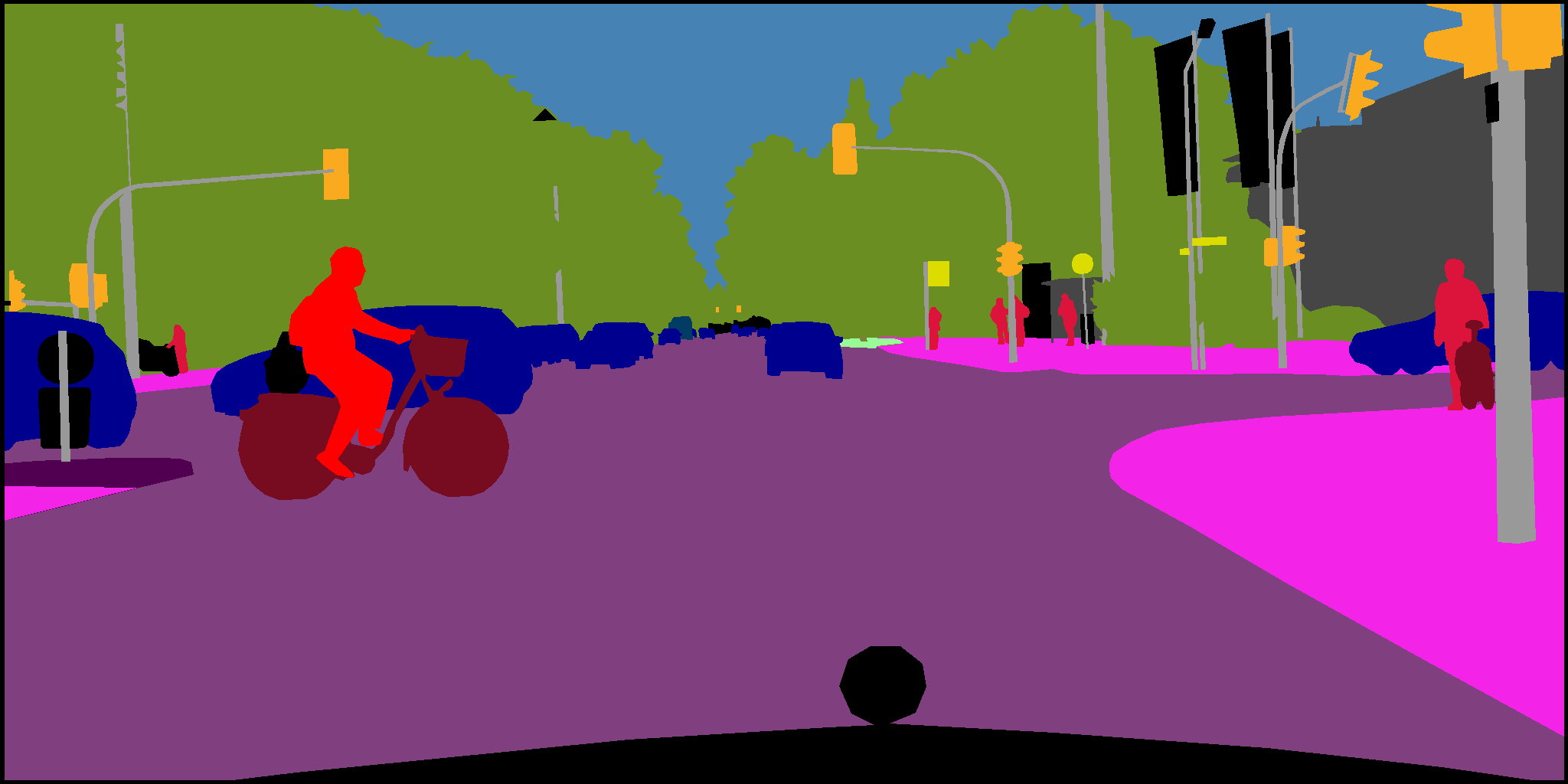} &
			\includegraphics[width=0.19\linewidth]{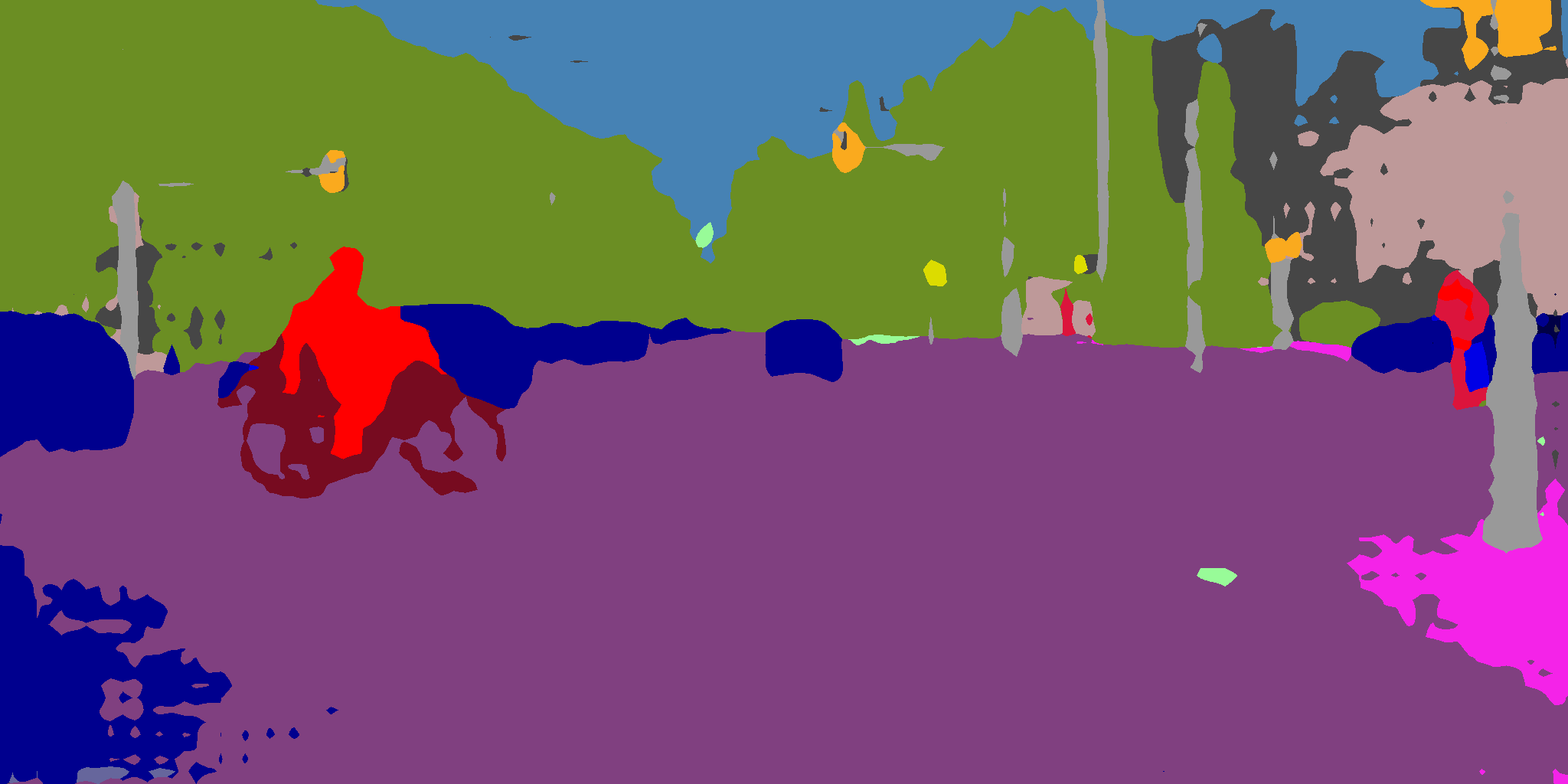} &
			\includegraphics[width=0.19\linewidth]{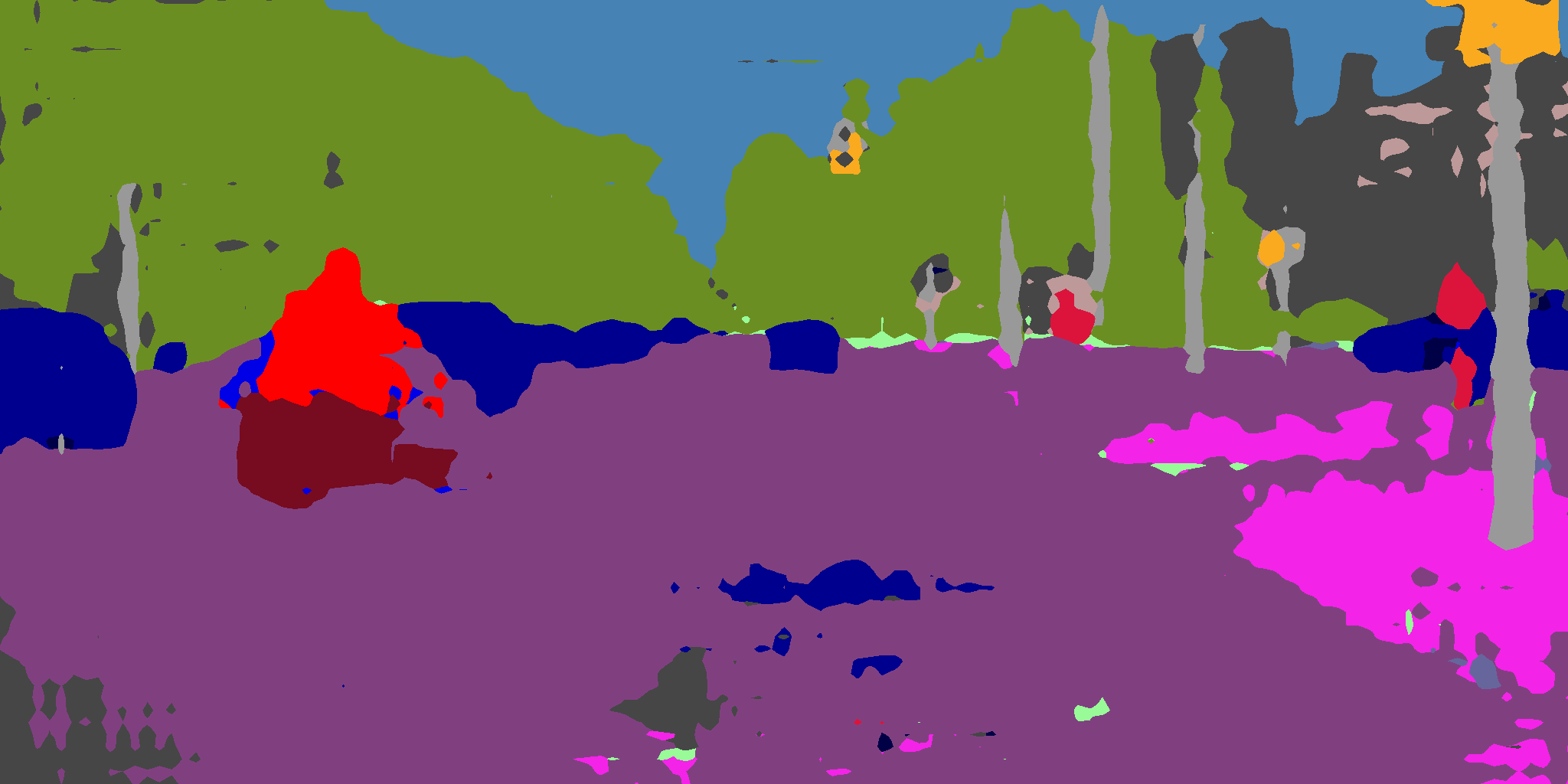} &
			\includegraphics[width=0.19\linewidth]{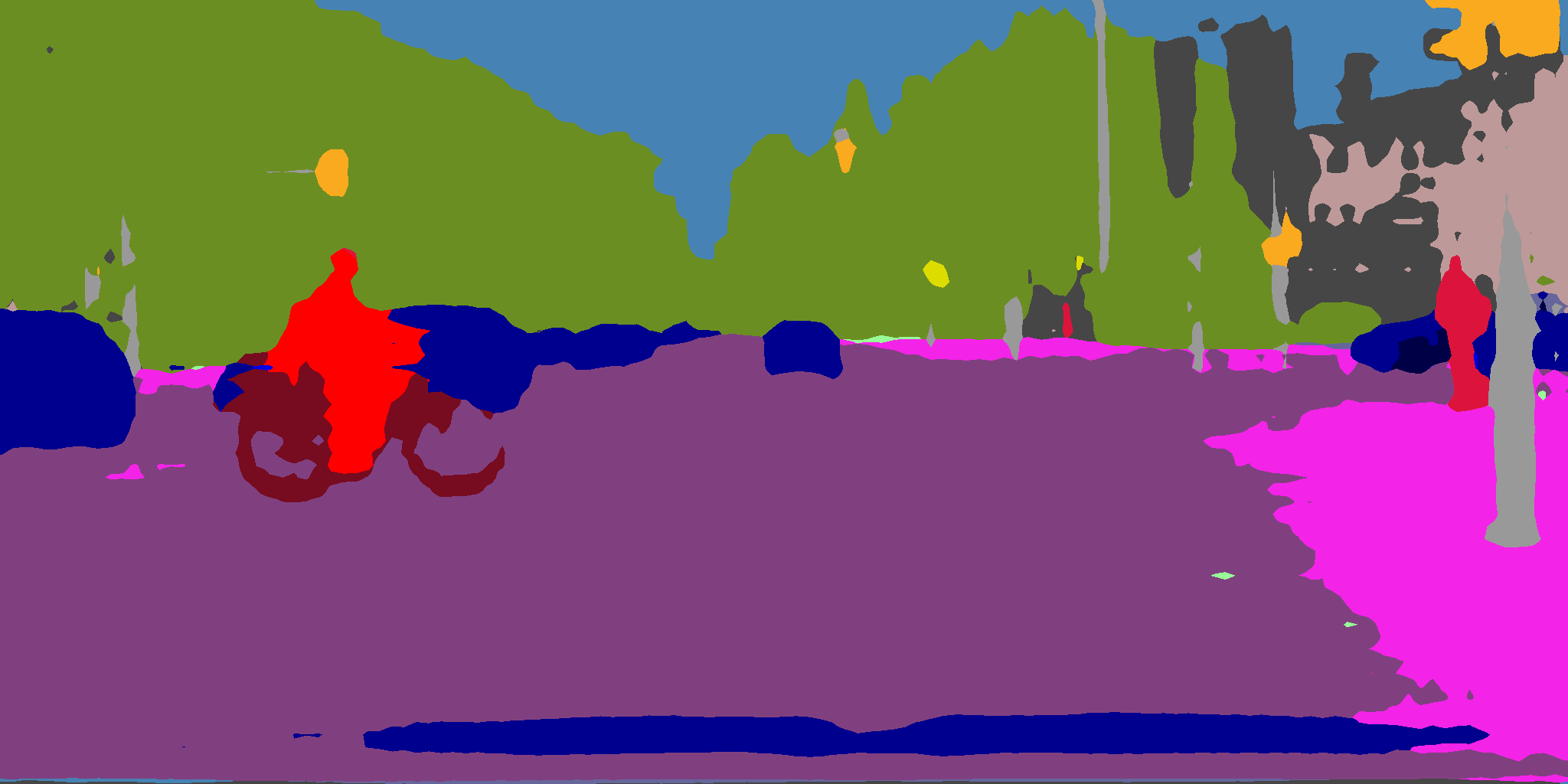} \\
			
			Target Image & Ground Truth & Before Adaptation & Feature Adaptation & Ours \\
			
		\end{tabular}
		\vspace{2mm}
		\caption{Example results of adapted segmentation for GTA5-to-Cityscapes. For each target image, we show results before adaptation, with feature adaptation and our adapted segmentations in the output space.
		}
		\label{fig:visual}
	\end{figure*}

On the other hand, we argue that utilizing a stronger baseline model is critical for understanding the importance of different adaptation components as well as for enhancing the performance to enable real-world applications.
Thus, we use the ResNet-101 based network introduced in Section \ref{sec:network} and train the proposed adaptation model.
%
	%
Table \ref{table:gta5_all} shows the baseline results only trained on source images without adaptation, with comparisons to our adapted models under different settings, including feature adaptation and single/multi-level adversarial learning in the output space.
Figure \ref{fig:visual} presents some example results for adapted segmentation.
We note that for small objects such as poles and traffic signs, they are harder to adapt since they easily get merged with background classes.

	In addition, another factor to evaluate the adaptation performance is to measure how much gap is narrowed between the adaptation model and the fully-supervised model.
Hence, we train the model using annotated ground truths in the Cityscapes dataset as the oracle results.
Table \ref{table:gta5_gap} shows the gap under different baseline models.
We observe that, although the oracle result does not differ a lot between VGG-16 and ResNet-101 based models, the gap is larger for the VGG one.
It suggests us that to narrow the gap, using a deeper model with larger capacity is more practical.
	%
\vspace{-3mm}
	{\flushleft {\bf Parameter Analysis.}}
	During optimizing the segmentation network $\mathbf{G}$, it is essential to balance the weight between segmentation and adversarial losses.
	We first consider the single-level case in \eqref{eq:loss} and conduct experiments to observe the impact of changing $\lambda_{adv}$.
	Table \ref{table:gta5_para} shows that a smaller $\lambda_{adv}$ may not
facilitate the training process significantly, while a larger $\lambda_{adv}$ may propagate incorrect gradients to the network.
We empirically choose $\lambda_{adv}$ as 0.001 in the single-level setting.
	%
	%
	\begin{table} [t]
		\caption{
			Sensitivity analysis of $\lambda_{adv}$ for feature/output space domain adaptation in the proposed method.
			We show that output space adaptation can tolerate a wide range of $\lambda_{adv}$, while it is sensitive to change $\lambda_{adv}$ for feature adaptation.
		}
		\vspace{1mm}
		\label{table:gta5_para}
		\small
		\centering
		\renewcommand{\arraystretch}{1.2}
		\setlength{\tabcolsep}{2.5pt}
		\begin{tabular}{lcccc}
			\toprule
			
			& \multicolumn{4}{c}{GTA5 $\rightarrow$ Cityscapes} \\
			\midrule
			
			$\lambda_{adv}$ & 0.0005 & 0.001 & 0.002 & 0.004 \\
			
			\midrule
			
			Feature & 35.3 & 39.3 & 35.9 & 32.8 \\
			Output Space & 40.2 & 41.4 & 40.4 & 40.1 \\
			
			\bottomrule
		\end{tabular}
		\vspace{-3mm}
	\end{table}
\vspace{-3mm}
{\flushleft {\bf Feature Level v.s. Output Space Adaptation.}}
	In the single-level setting in \eqref{eq:loss}, we compare results by using feature-level or output space adaptation via adversarial learning.
	For feature-level adaptation, we adopt a similar strategy as used in \cite{Hoffman_CoRR_2016, Chen_ICCV_2017} and train our model accordingly.
	Table \ref{table:gta5_all} shows that the proposed adaptation method in the output space performs better than the one in the feature level.

In addition, Table \ref{table:gta5_para} shows that adaptation in the feature space is more sensitive to $\lambda_{adv}$, which causes the training process more difficult, while output space adaptation allows for a wider range of $\lambda_{adv}$.
	%
One reason is that as feature adaptation is performed in the high-dimensional space, the problem for the discriminator becomes easier.
Thus, such an adapted model cannot effectively match distributions between source and target domains via adversarial learning.
	%
\vspace{-3mm}
{\flushleft {\bf Single-level v.s. Multi-level Adversarial Learning.}}
We have shown the merits of adopting adversarial learning in the output space.
In addition, we present the results of using multi-level adversarial learning in Table \ref{table:gta5_all}.
Here, we utilize an additional adversarial module (see Figure \ref{figure: framework}) and  
jointly optimize \eqref{eq:loss_multi} for two levels.
To properly balance $\lambda^i_{seg}$ and $\lambda^i_{adv}$, we use the same weight as in the single-level setting for the high-level output space (i.e., $\lambda^1_{seg}$ = 1 and $\lambda^1_{adv}$ = 0.001).
Since the low-level output carries less information to predict the segmentation, we use smaller weights for both the segmentation and adversarial loss (i.e., $\lambda^2_{seg}$ = 0.1 and $\lambda^2_{adv}$ = 0.0002).
Evaluation results show that our multi-level adversarial adaptation further improves the segmentation accuracy. 
More results and analysis are presented in the supplementary material.
	\begin{table*} [t]
		\caption{
	Results of adapting SYNTHIA to Cityscapes. We first compare our results using single-level adversarial learning in the output space with other state-of-the-art algorithms with the VGG-16 based model.
	Then we adopt the ResNet-101 based model and present ablation study on different components of our proposed method.			
	}
		\vspace{1mm}
		\label{table:synthia}
		\small
		\centering
		\renewcommand{\arraystretch}{1.0}
		\setlength{\tabcolsep}{6.5pt}
		\begin{tabular}{lcccccccccccccc}
			\toprule
			
			& \multicolumn{14}{c}{SYNTHIA $\rightarrow$ Cityscapes} \\
			\midrule
			
			Method & \rotatebox{90}{road} & \rotatebox{90}{sidewalk} & \rotatebox{90}{building} & \rotatebox{90}{light} & \rotatebox{90}{sign} & \rotatebox{90}{veg} & \rotatebox{90}{sky} & \rotatebox{90}{person} & \rotatebox{90}{rider} & \rotatebox{90}{car} & \rotatebox{90}{bus} & \rotatebox{90}{mbike} & \rotatebox{90}{bike} & mIoU\\
			
			\midrule
			
			FCNs in the Wild \cite{Hoffman_CoRR_2016} & 11.5 & 19.6 & 30.8 & 0.1 & \textbf{11.7} & 42.3 & 68.7 & \textbf{51.2} & 3.8 & 54.0 & 3.2 & 0.2 & 0.6 & 22.9 \\
			
			CDA \cite{Zhang_ICCV_2017} & 65.2 & 26.1 & 74.9 & \textbf{3.7} & 3.0 & 76.1 & 70.6 & 47.1 & 8.2 & 43.2 & \textbf{20.7} & 0.7 & \textbf{13.1} & 34.8 \\
			
			Cross-City \cite{Chen_ICCV_2017} & 62.7 & 25.6 & \textbf{78.3} & 1.2 & 5.4 & \textbf{81.3} & \textbf{81.0} & 37.4 & 6.4 & 63.5 & 16.1 & 1.2 & 4.6 & 35.7 \\
			
			Ours (single-level) & \textbf{78.9} & \textbf{29.2} & 75.5 & 0.1 & 4.8 & 72.6 & 76.7 & 43.4 & \textbf{8.8} & \textbf{71.1} & 16.0 & \textbf{3.6} & 8.4 & \textbf{37.6} \\
			
			\midrule
			Baseline (ResNet) & 55.6 & 23.8 & 74.6 & 6.1 & \textbf{12.1} & 74.8 & 79.0 & \textbf{55.3} & 19.1 & 39.6 & 23.3 & 13.7 & 25.0 & 38.6 \\
			
			Ours (feature) & 62.4 & 21.9 & 76.3 & \textbf{11.7} & 11.4 & 75.3 & 80.9 & 53.7 & 18.5 & 59.7 & 13.7 & 20.6 & 24.0 & 40.8 \\
			
			Ours (single-level)  & 79.2 & 37.2 & \textbf{78.8} & 9.9 & 10.5 & \textbf{78.2} & 80.5 & 53.5 & 19.6 & 67.0 & 29.5 & \textbf{21.6} & 31.3 & 45.9 \\
			
			Ours (multi-level) & \textbf{84.3} & \textbf{42.7} & 77.5 & 4.7 & 7.0 & 77.9 & \textbf{82.5} & 54.3 & \textbf{21.0} & \textbf{72.3} & \textbf{32.2} & 18.9 & \textbf{32.3} & \textbf{46.7} \\
			
			\bottomrule
		\end{tabular}
		\vspace{-2mm}
	\end{table*}
	\begin{table} [t]
		\caption{Performance gap between the adapted model and the fully-supervised (oracle) model.
			We first compare results with state-of-the-art methods using the VGG based model, and then show our result using the ResNet one.}
		\vspace{1mm}
		\label{table:synthia_gap}
		\small
		\centering
		\renewcommand{\arraystretch}{1.0}
		\setlength{\tabcolsep}{2.5pt}
		\begin{tabular}{lcccc}
			\toprule
			
			& \multicolumn{3}{c}{SYNTHIA $\rightarrow$ Cityscapes} \\
			\midrule
			
			Method & Baseline & Adapt & Oracle & mIoU Gap \\
			
			\midrule
			
			FCNs in the Wild \cite{Hoffman_CoRR_2016} & \multirow{4}{*}{VGG-16} & 22.9 & 73.8 & -50.9 \\
			
			CDA \cite{Zhang_ICCV_2017}  & & 34.8 & 69.6 & -34.8 \\ 
			
			Cross-City \cite{Chen_ICCV_2017} & & 35.7 & 73.8 & -38.1 \\
			
			Ours (single-level) & & 37.6 & 68.4 & -30.8 \\
			
			\midrule
			
			Ours (multi-level) & ResNet-101 & 46.7 & 71.7 & -25.0 \\
			\bottomrule
		\end{tabular}
	\vspace{-2mm}
	\end{table}
	\subsection{SYNTHIA}
To adapt from the SYNTHIA to Cityscapes datasets, we use the SYNTHIA-RAND-CITYSCAPES~\cite{Ros_CVPR_2016} set as the source domain which contains 9400 images compatible with the cityscapes annotated classes.
Similar to \cite{Chen_ICCV_2017}, we evaluate images 
on the Cityscapes validation set with 13 classes.
For the weight in \eqref{eq:loss} and \eqref{eq:loss_multi}, we use the same ones as in the case of GTA5 dataset.
	%
	
	
	%
Table \ref{table:synthia} shows evaluation results of the proposed algorithm 
against the state-of-the-art methods \cite{Chen_ICCV_2017, Hoffman_CoRR_2016, Zhang_ICCV_2017} that use feature adaptation.
Similar to the experiments with the GTA5 dataset, we first utilize the same VGG-based model and train our single-level adaptation model for fair comparisons.
The experimental results suggest that adapting the model in the output space performs better.
Second, we compare results using different components of the proposed method with the ResNet based model.
We show that the multi-level adaptation module improves the results over the baseline, feature space adaptation and single-level adaptation models.
In addition, we present comparisons of mean IoU gap between adapted and oracle results in Table \ref{table:synthia_gap}.
Our method achieves the smallest gap and is the only one that can minimize the gap below 30\%.
	%
%
%
	%
	\begin{table*} [t]
		\caption{
			Results of adapting Cityscapes to the Cross-City dataset.
			We construct our baseline model using the ResNet-101 architecture, and compare results between feature adaptation and our multi-level adaptation method in the output space.
		}
		\vspace{1mm}
		\label{table:cross-city}
		\small
		\centering
		\renewcommand{\arraystretch}{1.2}
		\setlength{\tabcolsep}{5.5pt}
		\begin{tabular}{llcccccccccccccc}
			\toprule
			
			& \multicolumn{14}{c}{Cityscapes $\rightarrow$ Cross-City} \\
			\midrule
			
			City & Method & \rotatebox{90}{road} & \rotatebox{90}{sidewalk} & \rotatebox{90}{building} & \rotatebox{90}{light} & \rotatebox{90}{sign} & \rotatebox{90}{veg} & \rotatebox{90}{sky} & \rotatebox{90}{person} & \rotatebox{90}{rider} & \rotatebox{90}{car} & \rotatebox{90}{bus} & \rotatebox{90}{mbike} & \rotatebox{90}{bike} & mIoU\\
			
			\midrule
			
			\multirow{4}{*}{Rome} & Cross-City \cite{Chen_ICCV_2017} & 79.5 & 29.3 & 84.5 & 0.0 & 22.2 & 80.6 & 82.8 & 29.5 & 13.0 & 71.7 & 37.5 & 25.9 & 1.0 & 42.9 \\
			
			& Our Baseline & \textbf{83.9} & \textbf{34.3} & 87.7 & 13.0 & \textbf{41.9} & 84.6 & 92.5 & 37.7 & \textbf{22.4} & 80.8 & 38.1 & 39.1 & 5.3 & 50.9  \\
			
			& Ours (feature) & 78.8 & 28.6 & 85.5 & 16.6 & 40.1 & 85.3 & 79.6 & 42.4 & 20.7 & 79.6 & \textbf{58.8} & 45.5 & 6.1 & 51.4 \\
			
			& Ours (output space) & \textbf{83.9} & 34.2 & \textbf{88.3} & \textbf{18.8} & 40.2 & \textbf{86.2} & \textbf{93.1} & \textbf{47.8} & 21.7 & \textbf{80.9} & 47.8 & \textbf{48.3} & \textbf{8.6} & \textbf{53.8} \\
			
			\midrule
			
			\multirow{4}{*}{Rio} & Cross-City \cite{Chen_ICCV_2017} & 74.2 & 43.9 & 79.0 & 2.4 & 7.5 & 77.8 & 69.5 & 39.3 & 10.3 & 67.9 & \textbf{41.2} & 27.9 & 10.9 & 42.5 \\
			
			& Our Baseline & \textbf{76.6} & \textbf{47.3} & 82.5 & \textbf{12.6} & 22.5 & 77.9 & 86.5 & 43.0 & 19.8 & \textbf{74.5} & 36.8 & 29.4 & 16.7 & 48.2 \\
			
			& Ours (feature) & 73.7 & 44.2 & 83.0 & 6.1 & 18.1 & 79.6 & 86.9 & 51.0 & 22.1 & 73.7 & 31.4 & \textbf{48.3} & \textbf{28.4} & 49.7 \\
			
			& Ours (output space) & 76.2 & 44.7 & \textbf{84.6} & 9.3 & \textbf{25.5} & \textbf{81.8} & \textbf{87.3} & \textbf{55.3} & \textbf{32.7} & 74.3 & 28.9 & 43.0 & 27.6 & \textbf{51.6} \\
			
			\midrule
			
			\multirow{4}{*}{Tokyo} & Cross-City \cite{Chen_ICCV_2017} & \textbf{83.4} & \textbf{35.4} & 72.8 & 12.3 & 12.7 & 77.4 & 64.3 & 42.7 & 21.5 & 64.1 & \textbf{20.8} & 8.9 & 40.3 & 42.8 \\
			
			& Our Baseline & 82.9 & 31.3 & \textbf{78.7} & 14.2 & 24.5 & 81.6 & 89.2 & 48.6 & 33.3 & 70.5 & 7.7 & 11.5 & 45.9 & 47.7 \\
			
			& Ours (feature) & 81.5 & 30.8 & 76.6 & 15.3 & 20.2 & 82.0 & 84.0 & 49.4 & 33.3 & 70.5 & 4.5 & 24.3 & \textbf{51.6} & 48.0 \\
			
			& Ours (output space) & 81.5 & 26.0 & 77.8 & \textbf{17.8} & \textbf{26.8} & \textbf{82.7} & \textbf{90.9} & \textbf{55.8} & \textbf{38.0} & \textbf{72.1} & 4.2 & \textbf{24.5} & 50.8 & \textbf{49.9} \\
			
			\midrule
			
			\multirow{4}{*}{Taipei} & Cross-City \cite{Chen_ICCV_2017} & 78.6 & 28.6 & 80.0 & 13.1 & 7.6 & 68.2 & 82.1 & 16.8 & 9.4 & 60.4 & 34.0 & 26.5 & 9.9 & 39.6 \\
			
			& Our Baseline & \textbf{83.5} & \textbf{33.4}& \textbf{86.6} & 12.7 & \textbf{16.4} & 77.0 & \textbf{92.1} & 17.6 & \textbf{13.7} & 70.7 & 37.7 & 44.4 & 18.5 & 46.5 \\
			
			& Ours (feature) & 82.1 & 31.9 & 84.1 & 25.7 & 13.2 & \textbf{77.2} & 81.2 & 28.1 & 12.0 & 67.0 & 35.8 & 43.5 & 20.9 & 46.6 \\
			
			& Ours (output space) & 81.7 & 29.5 & 85.2 & \textbf{26.4} & 15.6 & 76.7 & 91.7 & \textbf{31.0} & 12.5 & \textbf{71.5} & \textbf{41.1} & \textbf{47.3} & \textbf{27.7} & \textbf{49.1} \\
			
			\bottomrule
		\end{tabular}
	\end{table*}
	\subsection{Cross-City Dataset}

In addition to the synthetic-to-real adaptation for a larger domain gap, we conduct experiment on the Cross-City dataset~\cite{Chen_ICCV_2017} with smaller domain gaps between cities.
The dataset contains four different cities: Rio, Rome, Tokyo and Taipei, in which each city has 3200 images without annotations and 100 images with pixel-level ground truths for 13 classes.
Similar to~\cite{Chen_ICCV_2017}, we use the Cityscapes training set as the source domain and adapt it to each target city using 3200 images, while 100 annotated images are used for evaluation.
%
Since a smaller domain gap results in smaller output differences, we use smaller weights for the adversarial loss (i.e., $\lambda^i_{adv} = 0.0005$) when training our models, while the weights for segmentation remain the same as previous experiments.
	%
	
	%
We show our results in Table \ref{table:cross-city} with comparisons to \cite{Chen_ICCV_2017} and our baseline models under different settings.
Again, our final multi-level model achieves consistent improvement for different cities, which demonstrates the advantages of the proposed adaptation method in the output space.
%
Note that the state-of-the-art method \cite{Chen_ICCV_2017} uses a different baseline model, and we present it as a reference to analyze how much the proposed algorithm can improve.
%
\begin{table*} [!htb]
	\caption{Results of adapting GTA5 to Cityscapes.
	}
	\vspace{1mm}
	\label{table:gta5_ls}
	\footnotesize
	\centering
	\renewcommand{\arraystretch}{1.2}
	\setlength{\tabcolsep}{2.4pt}
	\begin{tabular}{lcccccccccccccccccccc}
		\toprule
		
		& \multicolumn{20}{c}{GTA5 $\rightarrow$ Cityscapes} \\
		\midrule
		
		Method & \rotatebox{90}{road} & \rotatebox{90}{sidewalk} & \rotatebox{90}{building} & \rotatebox{90}{wall} & \rotatebox{90}{fence} & \rotatebox{90}{pole} & \rotatebox{90}{light} & \rotatebox{90}{sign} & \rotatebox{90}{veg} & \rotatebox{90}{terrain} & \rotatebox{90}{sky} & \rotatebox{90}{person} & \rotatebox{90}{rider} & \rotatebox{90}{car} & \rotatebox{90}{truck} & \rotatebox{90}{bus} & \rotatebox{90}{train} & \rotatebox{90}{mbike} & \rotatebox{90}{bike} & mIoU\\
		
		\midrule
		
		Vanilla-GAN & 86.5 & 25.9 & 79.8 & 22.1 & 20.0 & 23.6 & 33.1 & \textbf{21.8} & 81.8 & 25.9 & 75.9 & 57.3 & 26.2 & 76.3 & 29.8 & 32.1 & \textbf{7.2} & \textbf{29.5} & \textbf{32.5} & 41.4 \\
		
		
		LS-GAN & \textbf{91.4} & \textbf{48.4} & \textbf{81.2} & \textbf{27.4} & \textbf{21.2} & \textbf{31.2} & \textbf{35.3} & 16.1 & \textbf{84.1} & \textbf{32.5} & \textbf{78.2} & \textbf{57.7} & \textbf{28.2} & \textbf{85.9} & \textbf{33.8} & \textbf{43.5} & 0.2 & 23.9 & 16.9 & \textbf{44.1} \\
		\bottomrule
	\end{tabular}
	\vspace{3mm}
\end{table*}
\begin{table*} [!htb]
	\caption{
		Results of adapting SYNTHIA to Cityscapes. mIoU and mIoU$^\ast$ are averaged over 16 and 13 categories, respectively.
	}
	\vspace{1mm}
	\label{table:synthia_ls}
	\footnotesize
	\centering
	\renewcommand{\arraystretch}{1.2}
	\setlength{\tabcolsep}{3pt}
	\begin{tabular}{lcccccccccccccccccc}
		\toprule
		
		& \multicolumn{18}{c}{SYNTHIA $\rightarrow$ Cityscapes} \\
		\midrule
		
		Method & \rotatebox{90}{road} & \rotatebox{90}{sidewalk} & \rotatebox{90}{building} & \rotatebox{90}{wall} & \rotatebox{90}{fence} & \rotatebox{90}{pole} & \rotatebox{90}{light} & \rotatebox{90}{sign} & \rotatebox{90}{veg} & \rotatebox{90}{sky} & \rotatebox{90}{person} & \rotatebox{90}{rider} & \rotatebox{90}{car} & \rotatebox{90}{bus} & \rotatebox{90}{mbike} & \rotatebox{90}{bike} & mIoU & mIoU$^\ast$ \\
		
		\midrule
		
		Vanilla-GAN & 79.2 & 37.2 & 78.8 & \textbf{10.5} & \textbf{0.3} & \textbf{25.1} & \textbf{9.9} & \textbf{10.5} & 78.2 & 80.5 & 53.5 & 19.6 & 67.0 & 29.5 & \textbf{21.6} & 31.3 & 39.5 & 45.9 \\
		
		LS-GAN & \textbf{84.0} & \textbf{40.5} & \textbf{79.3} & 10.4 & 0.2 & 22.7 & 6.5 & 8.0 & \textbf{78.3} & \textbf{82.7} & \textbf{56.3} & \textbf{22.4} & \textbf{74.0} & \textbf{33.2} & 18.9 & \textbf{34.9} & \textbf{40.8} & \textbf{47.6} \\
		
		\bottomrule
	\end{tabular}
\end{table*}
\begin{table*} [!htb]
	\caption{Results of adapting Synscapes to Cityscapes.
	}
	\vspace{1mm}
	\label{table:synscapes_ls}
	\footnotesize
	\centering
	\renewcommand{\arraystretch}{1.2}
	\setlength{\tabcolsep}{2.4pt}
	\begin{tabular}{lcccccccccccccccccccc}
		\toprule
		
		& \multicolumn{20}{c}{Synscapes $\rightarrow$ Cityscapes} \\
		\midrule
		
		Method & \rotatebox{90}{road} & \rotatebox{90}{sidewalk} & \rotatebox{90}{building} & \rotatebox{90}{wall} & \rotatebox{90}{fence} & \rotatebox{90}{pole} & \rotatebox{90}{light} & \rotatebox{90}{sign} & \rotatebox{90}{veg} & \rotatebox{90}{terrain} & \rotatebox{90}{sky} & \rotatebox{90}{person} & \rotatebox{90}{rider} & \rotatebox{90}{car} & \rotatebox{90}{truck} & \rotatebox{90}{bus} & \rotatebox{90}{train} & \rotatebox{90}{mbike} & \rotatebox{90}{bike} & mIoU\\
		
		\midrule
		
		Without Adaptation & 81.8 & 40.6 & 76.1 & 23.3 & 16.8 & 36.9 & 36.8 & 40.1 & 83.0 & 34.8 & 84.9 & 59.9 & 37.7 & 78.5 & 20.4 & 20.5 & 7.8 & 27.3 & 52.5 & 45.3 \\
		
		Vanilla-GAN & \textbf{94.2} & \textbf{60.9} & \textbf{85.1} & 29.1 & 25.2 & 38.6 & \textbf{43.9} & 40.8 & \textbf{85.2} & 29.7 & 88.2 & \textbf{64.4} & 40.6 & 85.8 & \textbf{31.5} & 43.0 & 28.3 & \textbf{30.5} & \textbf{56.7} & 52.7 \\
		
		
		LS-GAN & \textbf{94.2} & 60.5 & 85.0 & \textbf{29.2} & \textbf{25.6} & \textbf{39.8} & 43.4 & \textbf{43.8} & \textbf{85.2} & \textbf{35.9} & \textbf{88.3} & 63.2 & \textbf{41.1} & \textbf{87.2} & 30.8 & \textbf{44.2} & \textbf{29.8} & 28.5 & 53.7 & \textbf{53.1} \\
		\bottomrule
	\end{tabular}
	\vspace{3mm}
\end{table*}
\section{Concluding Remarks}
	%
In this paper, we exploit the fact that segmentations are structured outputs and share many similarities between source and target domains.
We tackle the domain adaptation problem for semantic segmentation via adversarial learning in the output space.
To further enhance the adapted model, we construct a multi-level adversarial network to effectively perform output space domain adaptation at different feature levels.
	Experimental results show that the proposed method performs favorably against numerous baseline models and the state-of-the-art algorithms.
We hope that our proposed method can be a generic adaptation model for a wide range of pixel-level prediction tasks.

\vspace{-2mm}
{\flushleft {\bf Acknowledgments.}}
W.-C. Hung is supported in part by the NSF CAREER Grant \#1149783, gifts from Adobe and NVIDIA. We especially thank Yan-Ting Liu for helping experiments using LS-GAN and the Synscapes dataset, as in the appendix.

	{\small
		\bibliographystyle{ieee}
		\bibliography{mybib}
	}

\appendix
\section{Least Squares Objective}
To analyze the impact of different type of GANs in our framework,
we adopt the least-squares loss function in \cite{mao2017least} that claims to generate higher-quality results and perform more stably during GAN training. The loss for discriminator training, similar to \eqref{eq:loss_d}, can be written as:
\begin{align}
	\label{eq:lsgan_loss_d}
	\mathcal{L}_{d}^{LS}(P) = \sum_{h,w}{} z \left(\mathbf{D}(P)^{(h,w,1)} - 1\right)^2 \\ \notag
	+ (1-z) \left(\mathbf{D}(P)^{(h,w,0)}\right)^2,
\end{align}
where $z = 0$ if the sample is drawn from the target domain, and $z = 1$ for the sample from the source domain.
Similar to \eqref{eq:loss_adv}, the adversarial loss can be written as:
\begin{equation}
	\mathcal{L}_{adv}^{LS}(I_t) = \sum_{h,w}{} 
	\left(\mathbf{D}(P_t)^{(h,w,1)} - 1\right)^2.
	\label{eq:lsgan_loss_adv}
\end{equation}
We use the single-level adaptation network and the ResNet-101 backbone as in the main paper, and all the other details are the same.
Results on Cityscapes using GTA5 and SYNTHIA as the source domain are presented in Table \ref{table:gta5_ls} and Table \ref{table:synthia_ls}, respectively.
We compare the performance of the vanilla GAN (as in the main paper) and the least-squares (LS) GAN.
Both tables show that using the LS-GAN objective achieves a higher mean IoU.
\section{Synscapes}
The Synscapes dataset \cite{synscapes} is a photorealistic synthetic dataset for street scene parsing. It consists of $25,000$ RGB images at $1440 \times 720$ resolution. The ground truth annotation adopts the Cityscapes convention that contains 19 categories. To adapt from Synscapes to Cityscapes, we use the entire Synsacpes dataset as the source domain.
In Table \ref{table:synscapes_ls}, we show results without adaptation, with vanilla GAN, and LS-GAN, using the single-level adaptation network and the ResNet-101 backbone.

Since the domain gap between Cityscapes and Synscapes is smaller than the case using either GTA5 or SYNTHIA as the source domain, the performance without adaptation already achieves a mean IoU of 45.3$\%$. By further using output space adaptation, the vanilla and LS GAN objectives improve the results and perform competitively.

\end{document}